\pdfoutput=1

\documentclass[11pt]{article}
\usepackage[]{acl}

\usepackage{times}
\usepackage{latexsym}
\usepackage{caption}
\usepackage{subcaption}
\usepackage{amsmath}
\usepackage{amssymb}
\usepackage{amsfonts}
\usepackage{subfloat}
\usepackage{xcolor}
\usepackage{siunitx}

\usepackage[T1]{fontenc}

\usepackage[utf8]{inputenc}

\usepackage{microtype}

\usepackage{inconsolata}

\usepackage{graphicx}
\usepackage{todonotes}

\usepackage{microtype}
\usepackage{graphicx}
\usepackage{tabularx}
\usepackage{xspace}
\usepackage{multirow}
\usepackage{enumitem}
\usepackage{booktabs}
\usepackage{adjustbox}

\newcommand{\sref}[1]{\S\ref{#1}}
\newcommand{\fref}[1]{Figure~\ref{#1}}
\newcommand{\tref}[1]{Table~\ref{#1}}
\newcommand{\human}{\textsc{Human}\xspace}
\newcommand{\chatbot}{\textsc{Chatbot}\xspace}
\newcommand{\simulator}{\textsc{Simulator}\xspace}
\newcommand{\myparagraph}[1]{\paragraph{#1}}
\newcommand{\LlamaThreeOneEightB}{\textsc{Llama3.1-8B}}
\newcommand{\LlamaThreeOneSeventyB}{\textsc{Llama3.1-70B}}

\newcommand{\MixtralEightBySevenB}{\textsc{Mixtral-8x7B}}
\newcommand{\MistralLarge}{\textsc{Mistral-Large}}

\newcommand{\CommandR}{\textsc{Command-R}}
\newcommand{\textttauto}[1]{\texttt{\hyphenchar\font=`\-#1}}

\title{Real or Robotic? Assessing Whether LLMs Accurately Simulate Qualities of Human Responses in Dialogue}

\author{
Jonathan Ivey\thanks{All authors have equal contribution, order is randomized except senior author}\textsuperscript{1,2}\hspace{0.5cm} 
Shivani Kumar\textsuperscript{1}\hspace{0.5cm} 
Jiayu Liu\textsuperscript{1,3}\hspace{0.5cm} 
Hua Shen\textsuperscript{1}\hspace{0.5cm} 
Sushrita Rakshit\textsuperscript{1}\hspace{0.5cm} \\
\textbf{Rohan Raju}\textsuperscript{1}\hspace{0.5cm} 
\textbf{Haotian Zhang}\textsuperscript{1}\hspace{0.5cm} 
\textbf{Aparna Ananthasubramaniam}\textsuperscript{1}\hspace{0.5cm} \\
\textbf{Junghwan Kim}\textsuperscript{1}\hspace{0.5cm} 
\textbf{Bowen Yi}\textsuperscript{1}\hspace{0.5cm} 
\textbf{Dustin Wright}\textsuperscript{1,4}\hspace{0.5cm} \textbf{Abraham Israeli}\textsuperscript{1}\hspace{0.5cm} \\
\textbf{Anders Giovanni Møller}\textsuperscript{1,5}\hspace{0.5cm} \textbf{Lechen Zhang}\textsuperscript{1}\hspace{0.5cm} 
\textbf{David Jurgens}\textsuperscript{1} \\
 \textsuperscript{1}University of Michigan \hspace{0.2cm}  \textsuperscript{2}University of Arkansas \hspace{0.2cm}
 \textsuperscript{3}University of Illinois Urbana-Champaign\\
 \textsuperscript{4}University of Copenhagen \hspace{0.2cm}\textsuperscript{5}IT University of Copenhagen \\
  \texttt{jurgens@umich.edu}\\
}

\begin{document}
\maketitle
\begin{abstract}
Studying and building datasets for dialogue tasks is both expensive and time-consuming due to the need to recruit, train, and collect data from study participants.
In response, much recent work has sought to use large language models (LLMs) to simulate both human-human and human-LLM interactions, as they have been shown to generate convincingly human-like text in many settings. However, to what extent do LLM-based simulations \textit{actually} reflect human dialogues? In this work, we answer this question by generating a large-scale dataset of 100,000 paired LLM-LLM and human-LLM dialogues from the WildChat dataset and quantifying how well the LLM simulations align with their human counterparts. Overall, we find relatively low alignment between simulations and human interactions, demonstrating a systematic divergence along the multiple textual properties, including style and content. Further, in comparisons of English, Chinese, and Russian dialogues, we find that models perform similarly. Our results suggest that LLMs generally perform better when the human themself writes in a way that is more similar to the LLM's own style.

\end{abstract}

\section{Introduction}
\label{sec:intro}
\begin{figure}[t]
    \centering
    \includegraphics[width=\linewidth]{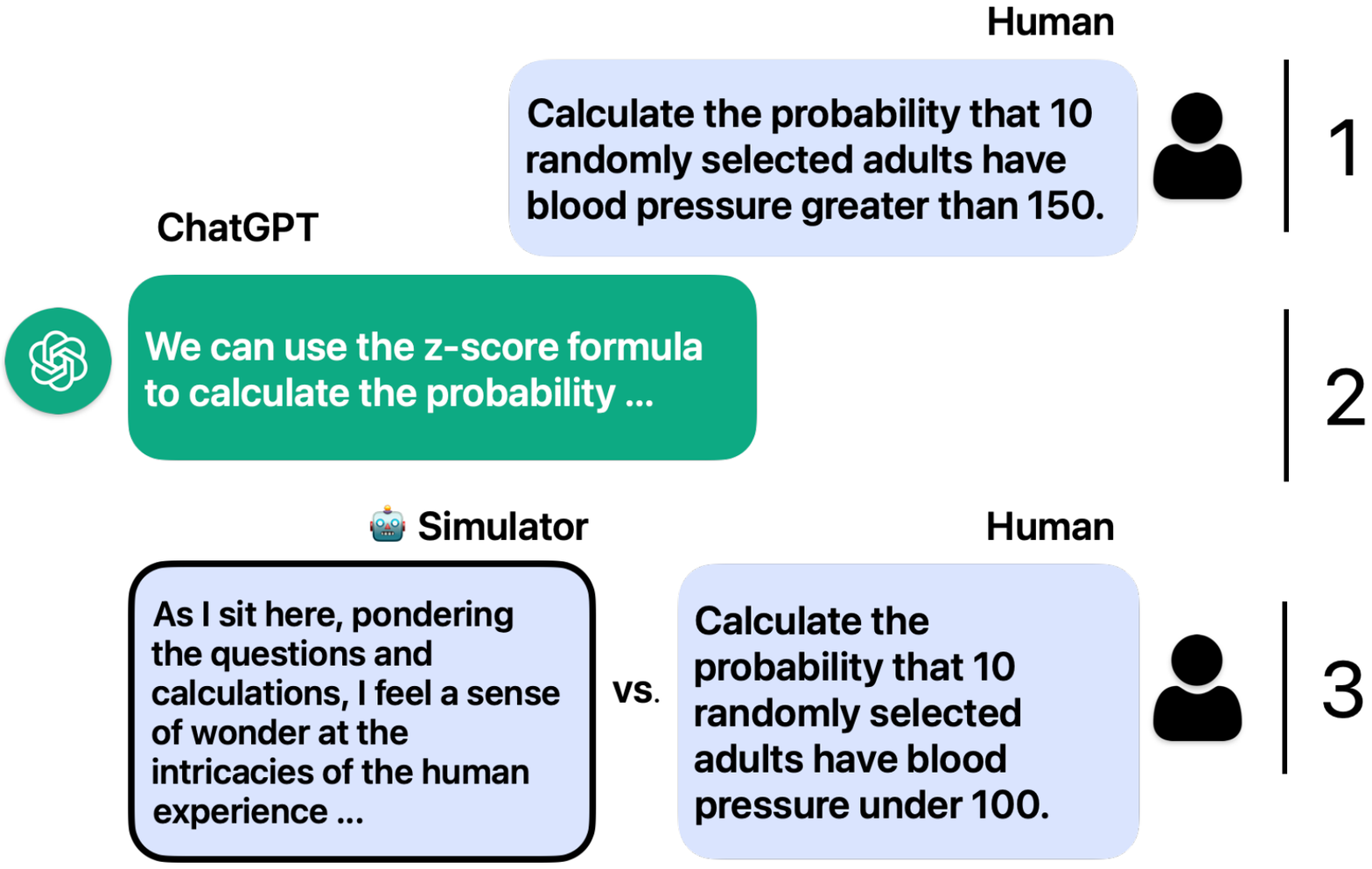}
    \caption{A sample conversation between a human and GPT-3.5 on WildChat and \LlamaThreeOneEightB's simulation of Turn 3 of the conversation.
    In this study, we compare the \simulator's output against the \human's output using 21 metrics, covering lexical, syntactic, semantic, and stylistic features.}
    \label{fig:sample_conversation}
\end{figure}

Large language models (LLMs) are capable of producing convincingly human-like responses to a broad range of inputs.
Recent work has explored the potential of LLMs to simulate human interactions in different scenarios \cite{zheng2023lmsys,kopf2024openassistant,zhao2024wildchat}.
The scenarios include simulating humans interacting with other humans to generate dialogue datasets, as well as simulating humans interacting with other LLMs \cite{Kim2024UnderstandingLM}, which is thought to be a scalable, automated approach for LLM quality testing.
Such simulations can greatly reduce the cost of collecting these data, which often require costly human labor and are difficult to scale to the diversity of LLM abilities.
However, this approach can only be effective if the responses generated by the LLM mirror how a human would interact in different scenarios.
In this work, we ask: to what extent can LLMs simulate the responses of humans in a human-LLM dialogue? %

To test simulation capabilities, we evaluate the degree to which LLMs mirror human behavior in real Human-LLM dialogues.
Our study asks the following three research questions:
(\textbf{RQ1}) to what extent does the choice of model and prompt instructions influence how well the LLM can simulate human behavior;
(\textbf{RQ2}) how do these results generalize to interactions in languages other than English; and
(\textbf{RQ3}) in what contexts are LLMs more likely to effectively simulate human responses?

To answer these questions, we develop an evaluation framework on top of human responses from the one million conversations in the WildChat dataset \citep{zhao2024wildchat}.
We compare the response from a human in a dialogue with the simulated response by an LLM, which we denote as a \simulator, as illustrated in Figure \ref{fig:sample_conversation}.
Responses are compared across multiple categories, e.g., lexicality, syntax, semantics, and style, to assess their fidelity to human behavior.
Using multiple LLMs and instruction prompts, we evaluate which settings led to better simulation; and, by using regression methods, we highlight the most significant factors that lead \simulator responses to be more human-like.

This paper makes the following four contributions.
First, we introduce a general evaluation framework for meaningful analysis of human-LLM simulations and a new dataset of over 1.2K annotator responses to the same conversations, providing a human-level performance comparison.
Second, we perform a large-scale analysis of 9 LLMs across 50 prompts simulating 2K English human-LLM conversations, and even the best model and prompt combinations are relatively weak at simulating human behavior.
Third, multilingual analyses on 10K Chinese and Russian human-LLM conversations show that performance is roughly similar, though still low.
Fourth, a regression analysis demonstrating which factors lead to more human-like responses from LLMs, we find that when the human begins the conversation in a writing style that resembles the LLM's, the \simulator can better match their behavior in a later turn.
All data and code are released for research purposes at {\small{\url{https://github.com/davidjurgens/human-llm-similarity}}}.

\section{Using LLMs to Simulate Human Interaction}
\label{sec:rel_works}

Given the practical use of LLMs in mimicking human turns in conversations, many studies deal with simulating human-LLM interactions and developing relevant conversational data sets for generic conversations
\cite{tamoyan2024llmroleplaysimulatinghumanchatbot, njifenjou2024roleplayzeroshotpromptinglarge, gosling2023pippapartiallysyntheticconversational} or for conversations of a specific domain, such as education \cite{abbasiantaeb2024let}, health \cite{cho2023can}, or programming \cite{liu2024make}.
Similarly, LLM-LLM interactions highlight the general capability of LLMs to mimic human discussions \cite{park2023generative, zhou2024real, rossetti2024social, zhou2023sotopia, chen2023places}.
In order to facilitate further research, a battery of studies have introduced various datasets of human-LLM \cite{zheng2023lmsys} and LLM-LLM \cite{kim2022soda, chen2023places} dialogue.
Another branch of interaction simulation has studied Digital Twins as virtual replicas of physical systems, in this case, humans in various discussion settings \cite{barricelli2019survey,barricelli2024digital,rossetti2024social,wen2024towards}.
It is common in many of these contexts for LLMs to augment---or even replace---human labor with simulations, thereby reducing time and effort \cite{kojima2022large, de2023leveraging}.

Humans, however, tend to have their specific style, intent, and self-creativity, which are challenging for LLMs to mimic despite recent technological breakthroughs \cite{stevenson2022putting,wu2023reasoning,wolf2023fundamental,gui2023challenge,jiang2024survey}.
Specifically, \citet{leng2023llm} highlight that while LLMs show promise for applications in social science research, further investigation into their subtle behavioral differences from humans and the development of robust evaluation protocols are essential, thus motivating our research.
Multiple metrics have been proposed to measure these differences, including content relevance \cite{abeysinghe2024challenges}, emotional alignment \cite{maehlum-etal-2024-difficult}, and intent accuracy \cite{Kim2024UnderstandingLM}.
A few studies \cite{sedoc2019chateval,svikhnushina2023approximating} have explored broader evaluation techniques by developing frameworks that align closely with human judgment.
However, comprehensive research about how different measures vary between LLMs and prompts is still lacking.
In this work, we address this measurement gap by analyzing how LLM responses compare to human responses across multiple similarity measures, LLM models, and prompts.

\section{Problem Definition}
\label{sec:prob_def}
Our research questions focus on understanding which settings lead to \simulator{s} producing convincingly \human-like dialogues.
We adopt the following modeling task, which provides a controlled setting for this question:
Given the initial human utterance in a conversation, followed by the response of a chatbot, we prompt a \simulator to suggest the next \human response in the discussion.
The ``true'' value of this third turn is known (but not given as input to the \simulator), as it is part of the WildChat dataset and thus acts as a reference response to which we compare.

This modeling setup offers the following desirable properties.
First, by using \human Turn 3 as a reference, we are able to investigate multiple factors related to the quality of LLM simulations: (i) the similarity between \simulator and \human responses along a variety of linguistic features, (ii) the influence of the original human Turn 1 and chatbot Turn 2 utterances, and (iii) the influence of model and prompt.
Second, limiting to \human Turn 3 allows us to study these factors \textit{in isolation} at \textit{the early stage of dialogue}, i.e., without the added influence of multiple (simulated or natural) turns.
This is desirable because the Turn 3 response demonstrates the degree to which the \simulator can continue a given conversation with the minimum amount of cues.
Finally, this setup maximizes the evaluation data set from WildChat while still containing multiple turns, as most conversations in the dataset end with 3 or fewer turns.

\section{Generating Simulation Data}
\label{sec:prompts}

To evaluate how well LLMs can be used to simulate human interactions, we generate a large dataset of dialogue simulations.
We start with the WildChat dialogues between \human{s} and a \chatbot (typically GPT-3.5) and then use another LLM as a \simulator to continue the conversation.

\myparagraph{Wildchat Data} %
\textit{WildChat} \cite{zhao2024wildchat} is a corpus of one million conversations between \human{s} and a \chatbot, comprising over 2.5 million interaction turns in multiple languages, with English accounting for 53\% of the turns.
We restrict ourselves to a sample of 102k English instances, along with 10k Chinese and 10k Russian instances for multilingual experiments in \sref{sec:exp2}.

To generate, we take the first two turns of each conversation (\human initiation and \chatbot response) as input to the \simulator and prompt the model to generate what the \human Turn 3 response would be, if any, or to indicate the human would not have responded.

\myparagraph{Prompt Composition}
Since the wording of a prompt can have a significant impact on an LLM's output and adopted persona~\cite{rottger2024political,DBLP:journals/corr/abs-2406-19238,DBLP:journals/corr/abs-2402-17649}, we conduct experiments with a wide range of prompts.
Working independently, 12 prompt writers familiar with LLMs composed a total of 50 candidate prompts for this task.
The prompt's aim is to have the \simulator match the conversational intent, content, and style of the first turn and to decide whether the conversation continues or not.
Prompt writers were given 10 randomly sampled dialogues from WildChat to guide their efforts in prompt generation.
They were encouraged to test their prompts using a \chatbot or other available tools.
Three example prompts are shown in Supplemental Table \ref{tab:selected-prompts}, and all 50 are in the Github repository.

\myparagraph{Prompt Classification}
Prompts generally took one of three strategies:
(1) \textbf{\textsc{cot}}: using chain-of-thought reasoning to guide the \simulator through its response;
(2) \textbf{\textsc{override}} employing various strategies to circumvent the \simulator's tendency to use overly polite, flowery, or long-winded language (e.g., jailbreaking, explicit instructions to ignore ethics training, jarring prompts);
and (3) \textbf{\textsc{direct}} directly instructing the \simulator to respond like a human, including telling the \simulator to pretend to be human or creating a persona for the model to follow (e.g., to-the-point, lazy).
Two authors jointly labeled each prompt by strategy, resulting in 13 prompts as \textsc{cot}, 14 as \textsc{override}, and 23 as \textsc{direct}.

\myparagraph{Generating and Parsing Simulated Turn 3}
Each of the 50 prompts was used to generate the third turn of a random sample of English-language dialogues from WildChat.
Nine models were used:
Meta-Llama-3.1-8B-Instruct~\citep{dubey2024llama3herdmodels}, Meta-Llama-3.1-70B-Instruct~\citep{dubey2024llama3herdmodels}, Meta-Llama-3-70B-Instruct~\citep{llama3modelcard}, Mistral-7B-Instruct-v0.3~\cite{jiang2023mistral7b}, Mixtral-8x7B-Instruct-v0.1~\citep{jiang2024mixtralexperts}, Mistral-Large-Instruct~\citep{mistrallarge}, Phi-3-medium-4k-instruct~\citep{abdin2024phi3technicalreporthighly}, Qwen2-72B-Instruct~\citep{qwen2}, and c4ai-command-r-v01~\citep{commandr}.
We use customized regular expressions for each prompt to parse and extract the \simulator's response text, especially for the 13 \textsc{cot} prompts.
After parsing, 4 prompts were discarded due to their frequent failure to produce valid output.
Details on inference are in Appendix \sref{sec:inferencedetail}.

\begin{figure*}[t]
    \centering
    \includegraphics[width=1.0\linewidth]{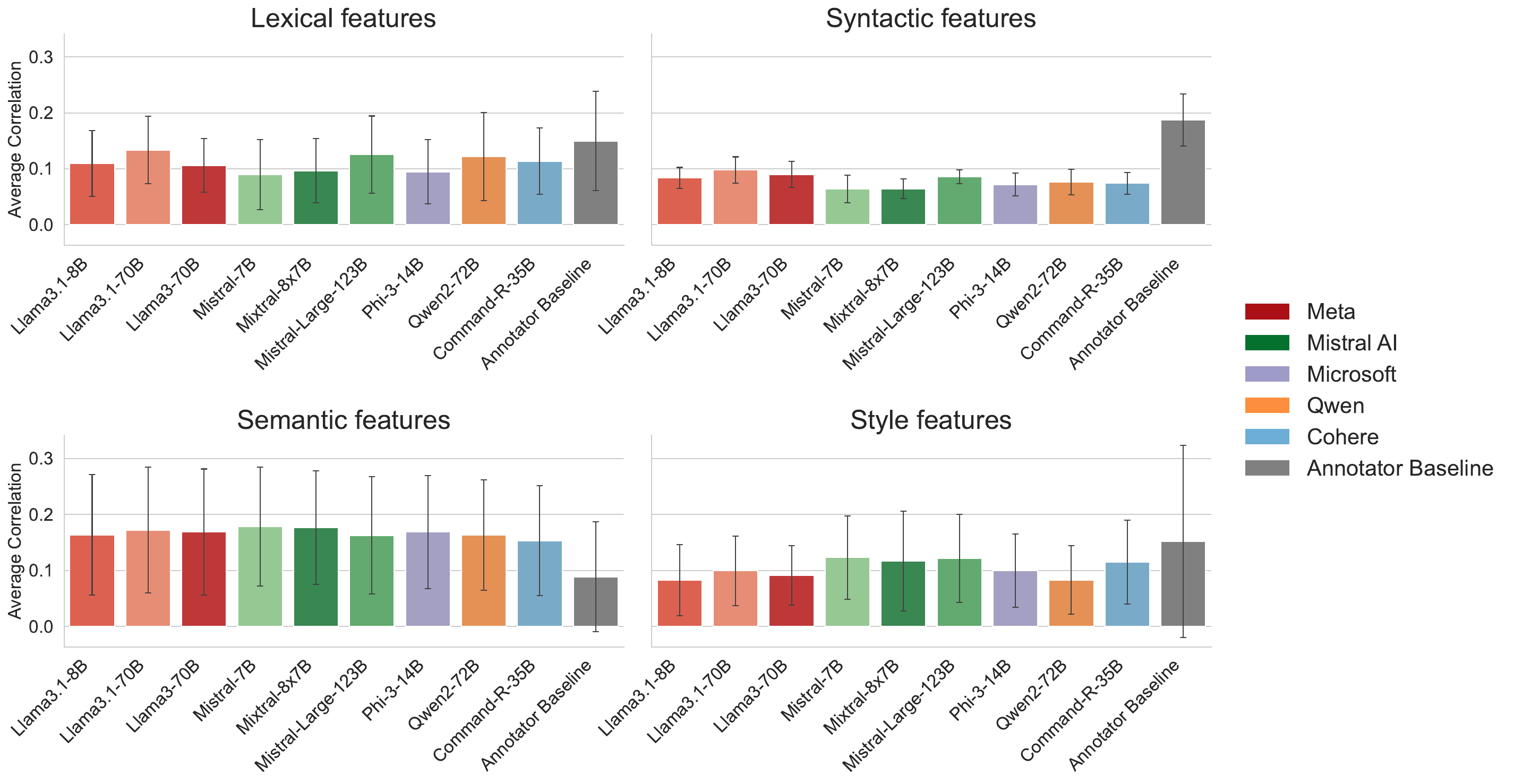}
    \caption{How well do LLMs simulate \human responses to a \chatbot? We compare the nine models used as \simulator{s} to the original \human by correlating properties of the text they write (\tref{tab:eval-metrics}).
    Bars represent the average correlation across all metrics in a category, and error bars are bootstrapped 95\% confidence intervals over these metrics.
    As a baseline, we also compare the performance of a human annotator on this task.
    There is limited cross-model variation in performance, and \simulator{s} tend to have higher performance in semantic features and lower performance in syntactic features, while the opposite is true of the human annotators.}
    \label{fig:model-differences}
\end{figure*}

\myparagraph{Evaluating LLM Simulations}
\label{sec:evaluating}
To evaluate, we select a breadth of 21 linguistic measures along 4 broad categories: style, lexical, syntactic, and semantic.
These categories and measures have been broadly used for different NLP tasks \cite{sebastiani2002machine,stamatatos2009survey,fu2018style,ribeiro2020beyond}.
The output of these measures can be either a scalar, a probability distribution, or a feature vector.
Scalars include both unbounded measures (e.g., utterance length) and bounded measures in $[0, 1]$ (e.g., text sentiment).
Probability distributions include, e.g., the probability that a token with a particular part of speech appears in a sentence across different parts of speech.
Finally, feature vectors include semantic (SBERT,~\citet{reimers-2019-sentence-bert}) and style (LUAR,~\citet{uar-emnlp2021}) embeddings.
In addition, we record if the \simulator and \human each end the conversation on Turn 3 as a binary measure.
A summary of all measures included in this study is given in \tref{tab:eval-metrics} in the Appendix.

\begin{table*}[ht]
    \centering
    \begin{tabular}{l|c|p{3.6in}}
        \textbf{Category} & \textbf{Measure} & \textbf{Description} \\\hline
         \textbf{End} 
                & F1 & Comparison of how often the \simulator ends the conversation whens the human ends it.\\
                \hline
         \textbf{Lexical} 
                & Utterance Length$^s$ & Log-transformed number of words.\\
                & Average Word Length$^s$ & Log-transformed number of characters per word.\\
                & Perplexity$^s$ & Log-transformed perplexity of the utterance, calculated using \texttt{lmppl} and \textsc{GPT-2} as model.\\
                & Typographical Errors$^s$ & Fraction of words that have typographical errors, counted using \texttt{pyspellchecker}.\\
                \hline
        \textbf{Syntactic} 
                & Part of Speech$^d$ & Distribution of the utterance's part of speech tags from \texttt{spaCy}.\\
                & Dependency Tree Depth$^s$ & Log-transformed depth of the dependency tree from \texttt{spaCy}.\\
                & Tree Breadth$^s$ & Log-transformed number of leaf nodes.\\
                & Tree Dependency Distance$^s$ & Log-transformed average distance between dependents.\\
                \hline
        \textbf{Semantic} 
                & SBERT$^v$ & Cosine similarity of utterance embeddings from the \texttt{all-MiniLM-L6-v2} language model \citep{reimers-2019-sentence-bert}.\\
                & LIWC$^d$ & Distribution of 69 LIWC categories from \citet{liwc}.\\ 
                & Prompt Type$^d$ & Distribution of categories from prompt classification tool \texttt{valpy/prompt-classification}.\\
                \hline
        \textbf{Style} 
                & Punctuation$^d$ & Distribution of punctuation characters.\\
                & Capitalization$^s$ & Fraction of letters that are capitalized.\\
                & Sentiment$^s$ & Distribution of positive, neutral, and negative sentiment from \texttt{distilbert-base-multilingual-cased-sentiments-student}.\\
                & Politeness$^s$ & From \texttt{Genius1237/xlm-roberta-large-tydip}~\citep{srinivasan-choi-2022-tydip}.\\
                & Formality$^s$ & From \texttt{s-nlp/mdeberta-base-formality-ranker}~\citep{dementieva-etal-2023-detecting}.\\
                & Toxicity$^s$ & Toxicity of tone and content, as judged by annotators \texttt{s-nlp/roberta\_toxicity\_classifier}.\\
                & Readability$^s$ & Distribution of Flesch reading ease scores \citep{45033}.\\
                & Subjectivity$^s$ & The average subjectivity score of words in the utterance from the sentiment polarity lexicon in \texttt{textblob}.\\
                & LUAR$^v$ & Author style embeddings using \texttt{rrivera1849/LUAR-CRUD} \citep{uar-emnlp2021}.\\
                \hline
    \end{tabular}
    \caption{Measures used to evaluate how well LLMs capture properties of human responses at Turn 3 of a conversation.
    Letter superscript indicates whether the difference between human and \simulator measurements are \textbf{(s)} scalar values (compared with l1-distance), \textbf{(d)} probability distributions (compared with Jensen-Shannon divergence), or \textbf{(v)} vector embeddings (compared with cosine distance).}
    \label{tab:eval-metrics}
\end{table*}

\section{Can LLMs Simulate Human Replies?}
\label{sec:exp1}

In the first experiment, we measure the overall similarity of \simulator{s} to \human{s} (\textbf{RQ1}) by assessing how simulation instructions influence \simulator similarity to \human messages across a variety of models and prompts.

\subsection{Experimental Setup}

For our first experiment, we use the setup from \sref{sec:prompts} to generate 828K simulated responses (9 models, 46 prompts, 2,000 English conversations).
To evaluate performance, we measure the similarity between the original \human Turn 3 responses and the \simulator Turn 3 responses in lexical, syntactic, semantic, and style domains.
Starting with the 21 metrics from \sref{sec:prompts}, we correlate each \human and \simulator Turn 3 metric.
For scalar metrics (e.g., length, classifier probabilities), similarity is the Pearson's R correlation between \human and \simulator values.
For distribution metrics (e.g., POS tags, topics), similarity is the average correlation of \human and \simulator class frequencies over all classes in the distribution.
For vector metrics (e.g., BERT, LUAR), similarity is the average correlation of each dimension of the \human and \simulator embedding;\footnote{A more standard approach for calculating similarity would be to take the average cosine similarity for each \human/\simulator pair. However, we chose to use correlations so the feature vector similarity scores would be comparable to other metrics. Doing so yields similarity scores that are highly correlated with the corresponding cosine similarities.}
since the dimensions of embeddings are not inherently meaningful, we first rotate the embeddings using principal components analysis, so each dimension corresponds to an embedding's alignment with the direction of a principal component.
We then average the correlations for all metrics within each category given in the first column of \tref{tab:eval-metrics} to create four category-wise similarity scores corresponding to lexical, syntactic, semantic, and style.
To measure the similarity in ending a conversation, we use the binary F1 score between whether the \human and \simulator decide to end the conversation.

Although we have a human ground truth for how the conversation did continue, to estimate the task's difficulty, we also had a separate group of 12 humans who did not author any WildChat data to perform the same task as the \simulator{s} to produce Turn 3 responses.
1273 turns were annotated, and we compared \simulator performance on the subset of turns where both the annotators and \simulator continued the conversation.
Full annotation details are in Appendix \sref{app:annotation}.

\subsection{Results}

\begin{figure}[t]
    \centering
    \includegraphics[width=1.0\linewidth]{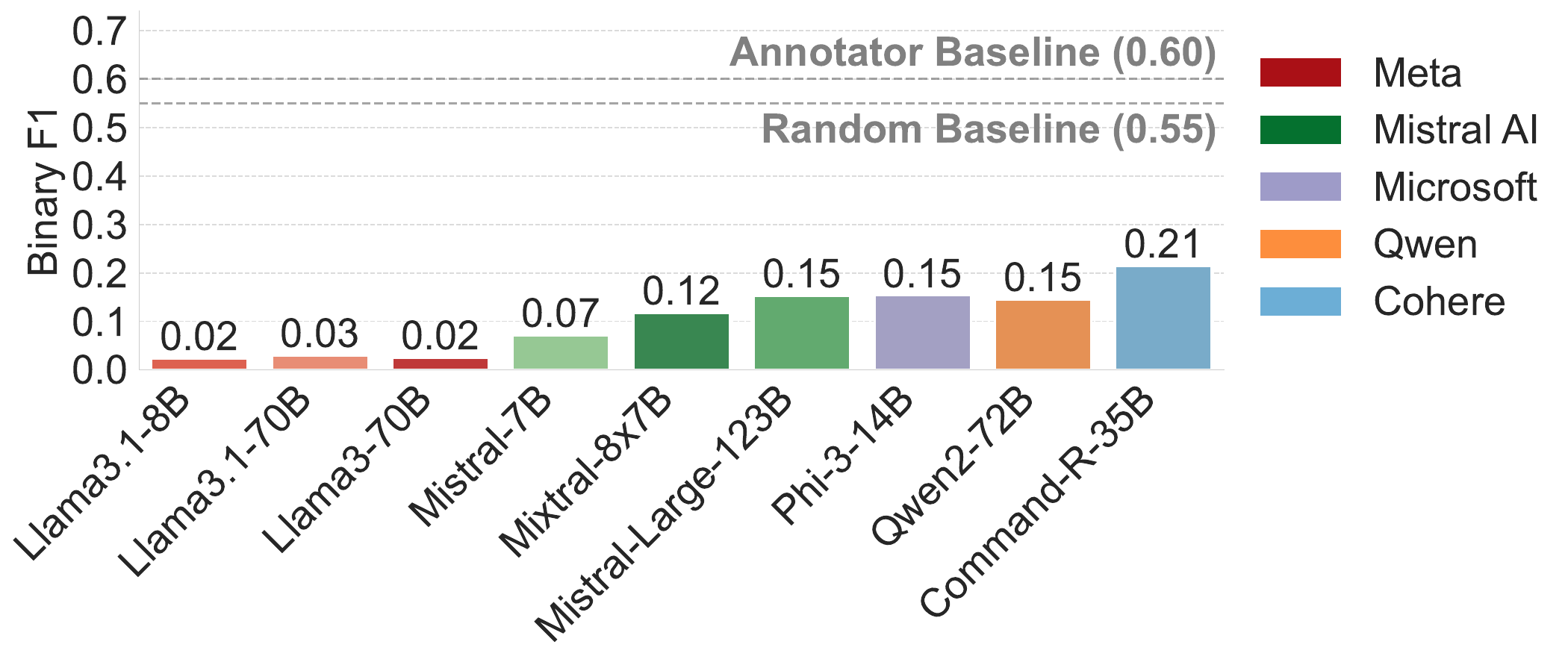}
    \caption{How well do LLMs predict whether \human{s} end a conversation with \chatbot after the first turn? Each bar represents the binary F1 score of each model predicting whether a conversation will end.
    The gray horizontal lines show the performance of human annotators and a random baseline that ends the conversation 50\% of the time.
    While there is inter-model variation, all models perform worse than chance.
    The human annotator performs better than chance.}
    \label{fig:end-differences}
\end{figure}

We compare the similarity of different \simulator{s} to the original \human in \fref{fig:model-differences}.
In general, similarity is lower across all settings for the syntactic measures, while semantic measures tend to be more similar.
Additionally, there is little variation across different \simulator{s}.
There is effectively no difference between the similarity of \simulator{s} and the similarity of the human annotator baseline across the lexical, semantic, and style categories of measures.
However, there is a discernible difference in how similar annotator utterances are syntactically to the original \human{s} compared to the \simulator{s}.
Additionally, for both the semantic and style categories, the confidence intervals for the annotator baseline extend beyond $0$, the performance we would expect from a totally random baseline, while each \simulator maintains non-random performance.
This finding suggests that humans and LLMs have complementary strengths in simulating dialogue; in order to most accurately reflect human utterances, a human-in-the-loop approach where \simulator{s} and annotators play complementary roles may be appropriate.
The correlation results for all individual metrics across \simulator{s} are shown in the Appendix in \tref{tab:ind_corr_models_en}.

In \fref{fig:end-differences}, we compare the similarity between \simulator{s} and \human{s} in their tendency to end the conversation.
\simulator{s} seldom end the conversation, continuing 87.1\% of the time for \CommandR\ to 98.6\% for \LlamaThreeOneEightB.
In contrast, human annotators more accurately mirror the original \human behavior in ending the conversation, indicating that collaboration between \simulator{s} and humans is effective for simulating human interactions.
Additionally, human annotators show a better ability to determine when a conversation is likely to end, more often predicting the end at Turn 2 when it actually occurs, whereas \simulator{s} tend to predict conversation end with similar frequency regardless of the \human{s} actual behavior (\fref{fig:f1-comparison}).
The performance differences across \simulator models are mainly driven by how frequently each model ends the conversation.

\begin{figure}[t]
    \centering
    \includegraphics[width=0.8\linewidth]{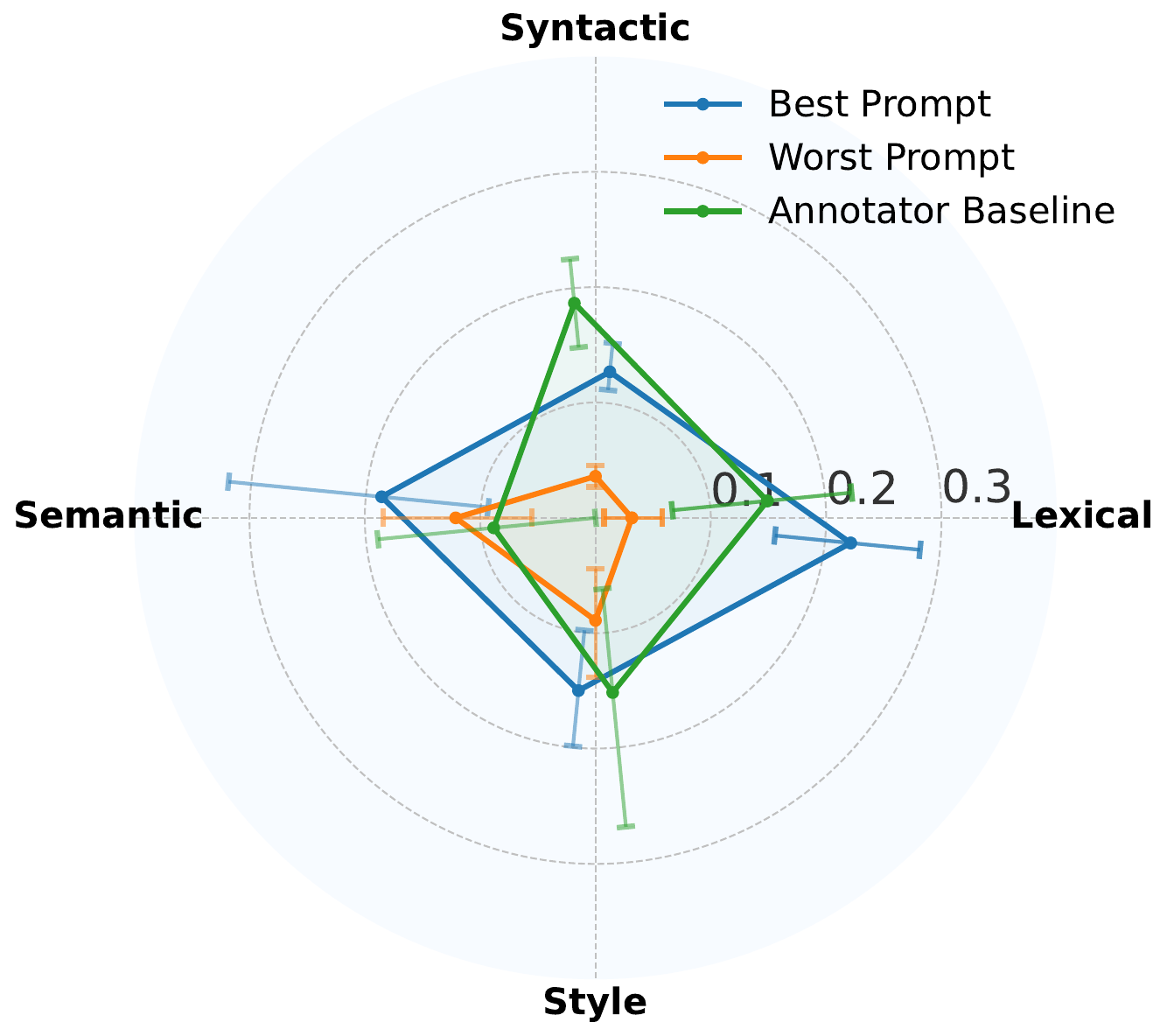}
    \caption{Using the methods from \fref{fig:model-differences}, the performance of the best and worst prompts and annotators are compared across metric categories.
    The best (a \textsc{direct} prompt) and worst (an \textsc{override} prompt) prompts are selected based on an overall average across all metrics and shown in \tref{tab:top-prompts}.
    The worst prompt underperforms the best prompt in all categories, and annotators outperform all prompts in syntax metrics.}
    \label{fig:prompt-differences}
\end{figure}

Finally, we look at the impact of prompt on \simulator and \human similarity in \fref{fig:prompt-differences}.
We plot the average similarity within each category for the best prompt (i.e., the highest average correlation across all metrics), worst prompt, and human annotator baseline.
See prompts in~\tref{tab:top-prompts}.
The best prompt shows higher lexical and semantic similarity than the human annotator baseline, while human annotators have higher syntactic similarity, in line with the results on models in \fref{fig:model-differences}.
We also see that prompts significantly affect the similarity across all categories, suggesting that prompt engineering impacts more than model selection for simulating human interactions.
This effect is robust based on the error bars of each method.
The most impacted categories are the lexical and syntactic categories, while the prompt has less effect on semantic and stylistic similarity.
Given that models also demonstrate little difference in these categories, it is challenging to engineer human simulations using LLMs that are semantically or stylistically similar, while it is possible with good prompt engineering to improve lexical and syntactic similarity.
The correlation results of all metrics across prompts are shown in Appendix \tref{tab:ind_corr_prompts_en}.

Overall, we find that the choice of \simulator has only a minor impact on the ability to simulate \human{s}, while the design of the prompt is most relevant when optimizing performance (\textbf{RQ1}).

\section{Simulation in non-English Languages}
\label{sec:exp2}

Next, we address how the results from \sref{sec:exp1} generalize to languages beyond English (\textbf{RQ2}).

\subsection{Experimental Setup}

We compare the performance of \simulator{s} across three languages: English, Chinese, and Russian.
Chinese and Russian were selected for their substantial representation in the WildChat dataset, following English.
Specifically, Chinese and Russian comprise 15.9\% and 10.5\%, respectively, whereas English accounts for 53\%.
For our analysis, we randomly sample $10,000$ conversations from each language.

To generate conversations, we use a subset of models from the previous section: \MistralLarge-123B, \LlamaThreeOneSeventyB, and \MixtralEightBySevenB.
We identify the most effective English prompts from each prompt category and have these prompts manually translated into Chinese and Russian by native speakers.
We select three prompts that consistently perform well across six largely uncorrelated metrics: capitalization, punctuation, part of speech, SBERT embeddings, sentiment, and politeness.(Appendix \sref{app:results}, \fref{fig:compare-prompts}).
The selected prompts are presented in Appendix \tref{tab:selected-prompts}.

To measure the similarity between \human and \simulator responses, we use a procedure similar to the one described in \sref{sec:exp1}.
Due to the unavailability of several metrics from \sref{sec:exp1} for Chinese and Russian, we focus on a subset of ten metrics that cover all four categories.
Detailed descriptions of these metrics and their operationalizations can be found in Appendix \tref{tab:eval-metrics-multilingual}.
We employ consistent or similarly trained models whenever possible to ensure comparability across languages.

\subsection{Results}

\begin{figure}[t]
    \centering
    \includegraphics[width=\linewidth]{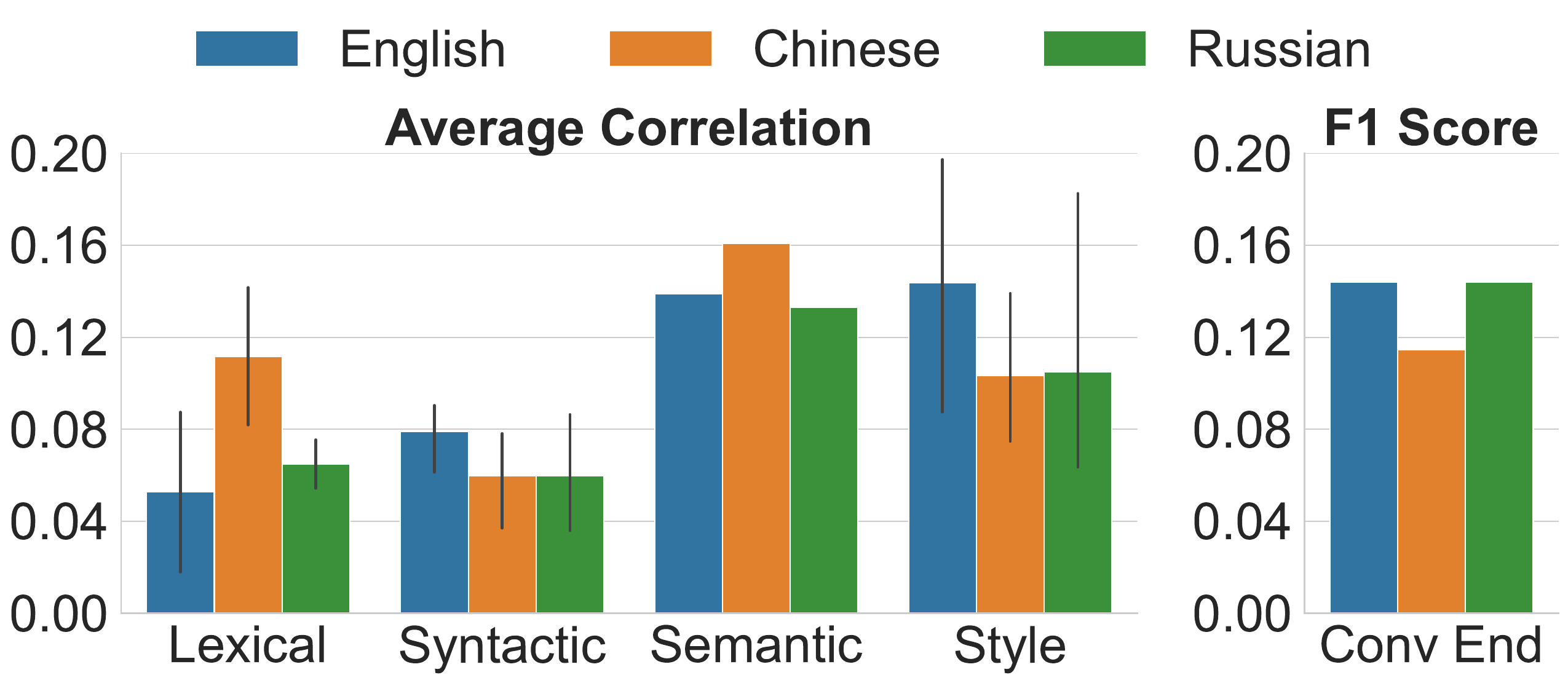}
    \caption{How well do \simulator{s} replicate \human text across languages? Similar to Figures~\ref{fig:model-differences} and \ref{fig:end-differences}, we plot the similarity between \simulator and \human text across ten metrics in three languages.
    English, Chinese, and Russian have similar performance patterns across all five categories of metrics.
    However, some differences exist (e.g., Chinese \simulator{s} outperform other languages in lexical and semantic metrics but underperform in conversation endings).
    Correlations of individual metrics are shown in Tables~\ref{tab:ind_corr_cn} and \ref{tab:ind_corr_ru}.}
    \label{fig:exp2}
\end{figure}

The aggregated metrics for the three languages are depicted in \fref{fig:exp2}.
Consistent with the findings discussed in \sref{sec:exp1}, all three languages show higher correlations between \simulator{s} and \human{s} in the semantic metric but lower correlations in the syntactic metric.
For detailed correlations of individual metrics, we refer to Appendix \tref{tab:ind_corr_cn} and \tref{tab:ind_corr_ru}.
Additionally, predicting whether a conversation will end consistently performs below chance across all languages, as \simulator{s} are unlikely to predict a conversation will end.
These observations reinforce the conclusions drawn in \sref{sec:exp1}.

\begin{figure*}[t]
    \centering
    \includegraphics[width=1.0\linewidth]{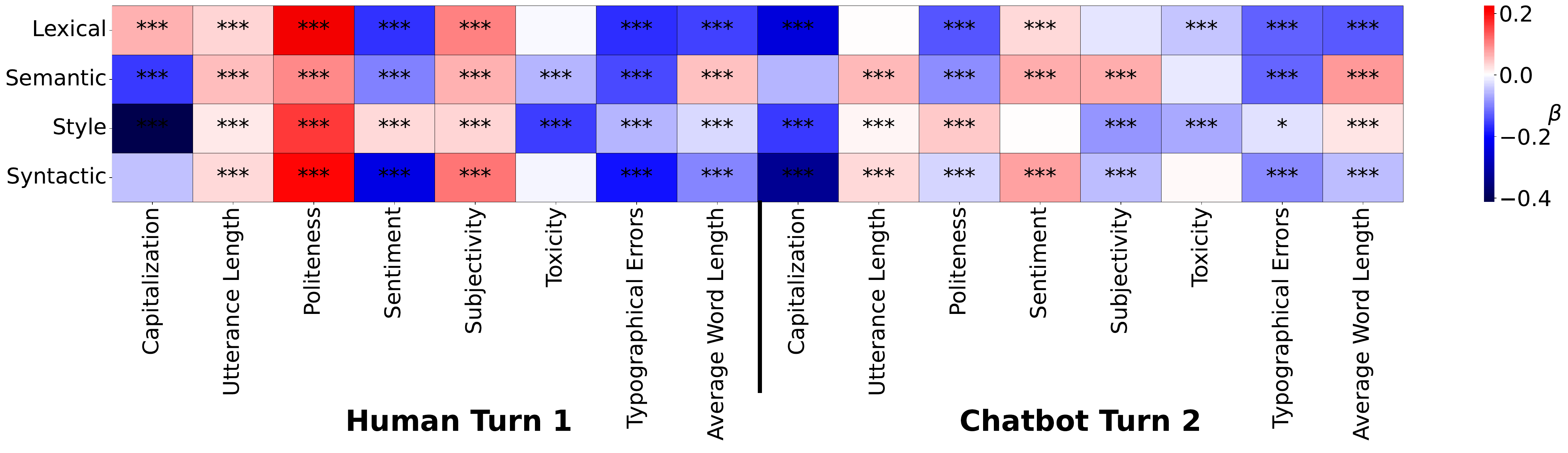}
    \caption{
    In what contexts do \simulator{s} best predict \human responses?
    We show the results of four regressions predicting the similarity between \simulator and \human at Turn 3 for different categories (rows), using \human Turn 1 and \chatbot Turn 2 properties as features (columns).
    We highlight a subset of the coefficients here, where red and blue colors indicate positive and negative regression coefficients $\beta$ respectively, and stars in each cell indicate the statistical significance of each $\beta$ after applying a Bonferroni correction (* p<0.05, ** p<0.01, *** p<0.001); Full regression coefficients are given in \tref{tab:regression-coefficients-all} and \tref{tab:regression-coefficients-all-continued}.
    The linguistic properties of \human Turn 1 have stronger effects than those of the \chatbot in Turn 2, showing that \simulator{s} do correctly accommodate more to the \human style.
    However, \simulator{s} tend to perform better when the \human{'s} Turn 1 more closely matches the properties of typical \simulator-generated text (e.g., more polite, fewer typos).
    }
    \label{fig:regression}
\end{figure*}

Notably, there are differences between the languages.
Chinese \simulator{s}, for example, outperform their English and Russian counterparts in lexical metrics and show slightly better performance in semantic metrics.
The differences in lexical metrics, such as utterance length and perplexity, may be attributed to the typically shorter sentence lengths in Chinese.
Conversely, English \simulator{s} excel in predicting style metrics compared to those in other languages.
Toxicity and sentiment metrics primarily contribute to this improvement.
In English, \simulator{s} more accurately reflect the variations in toxicity and sentiment of \human{s}.
This capability may vary across languages because Chinese and Russian have relatively smaller amounts of training data.
Consequently, safety training may impact these languages' outputs more significantly than English, leading to a strong prior on toxicity and sentiment that hinders the style match to \human{s}.
Additionally, English \simulator{s} demonstrate superior accuracy in modeling syntactic metrics.

The choice of model and prompting strategy affects performance, as shown in Appendix \fref{fig:ml-prompt-differences}.
The differences between models become more pronounced in languages other than English, as shown in \fref{fig:ml-models-differences}.
For example, while model differences in English are often minor, the smallest model (\MixtralEightBySevenB) frequently underperforms compared to other models in Chinese and Russian.
Moreover, the significant inter-model variation in Chinese and Russian may be due to larger models having more non-English samples in their pre-training data, which enhances their performance in non-English languages.
Additionally, the F1 score shows more variation across the models and prompting strategies than the text properties metrics, suggesting that engineering decisions may be more salient for closed-ended conversation-ending tasks than open-ended text generation tasks.
Appendix \sref{app:results} details the differences.

Overall, we find that LLMs' performance as \simulator{s} is largely consistent across English, Chinese, and Russian, although some differences suggest that languages for which \simulator{s} had less training data may have less robust performance across contexts (\textbf{RQ2}).

\section{When Do LLMs Successfully Simulate?}
\label{sec:exp3}

Experiments from \sref{sec:exp1} and \sref{sec:exp2} paint a high-level picture of how the \simulator{s} differ from \human{s} across settings and linguistic metrics.
However, an important question remains: \textit{when} do LLMs succeed as \simulator{s} of human conversations?
In our final experiment, we answer \textbf{RQ3} by investigating which factors directly impact the differences between \simulator{s} and \human{s}.

\subsection{Experimental Setup}

For this experiment, we focus on depth as opposed to breadth and generate Turn 3 utterances for a random sample of 100,000 conversations from the English-language WildChat corpus.
As in \sref{sec:exp2}, we generate responses using the best-performing English prompts from \sref{sec:exp1} with \MistralLarge-123B, \LlamaThreeOneSeventyB, and \MixtralEightBySevenB.
After generating responses, we perform a regression analysis to identify factors linked with higher or lower similarity between \simulator and \human responses.
For regression, we use the average similarity across the measures in \tref{tab:eval-metrics} as the dependent variable.
For independent variables, we use the conversation metadata (the \chatbot model used, the region where the \human is located, the \simulator model used, and the prompt), a subset of scalar metrics run on \human Turn 1 and \chatbot Turn 2 utterances (capitalization, utterance length, politeness, sentiment, subjectivity, toxicity, typographical errors, and average word length), and a set of 50 topics generated using Latent Dirichlet Analysis (LDA) on \human Turn 1 and \chatbot Turn 2 utterances.
Further details on the regression are given in Appendix \ref{app:regression_details}.

\subsection{Results}

\begin{figure}[t]
    \centering
    \includegraphics[width=1.0\linewidth]{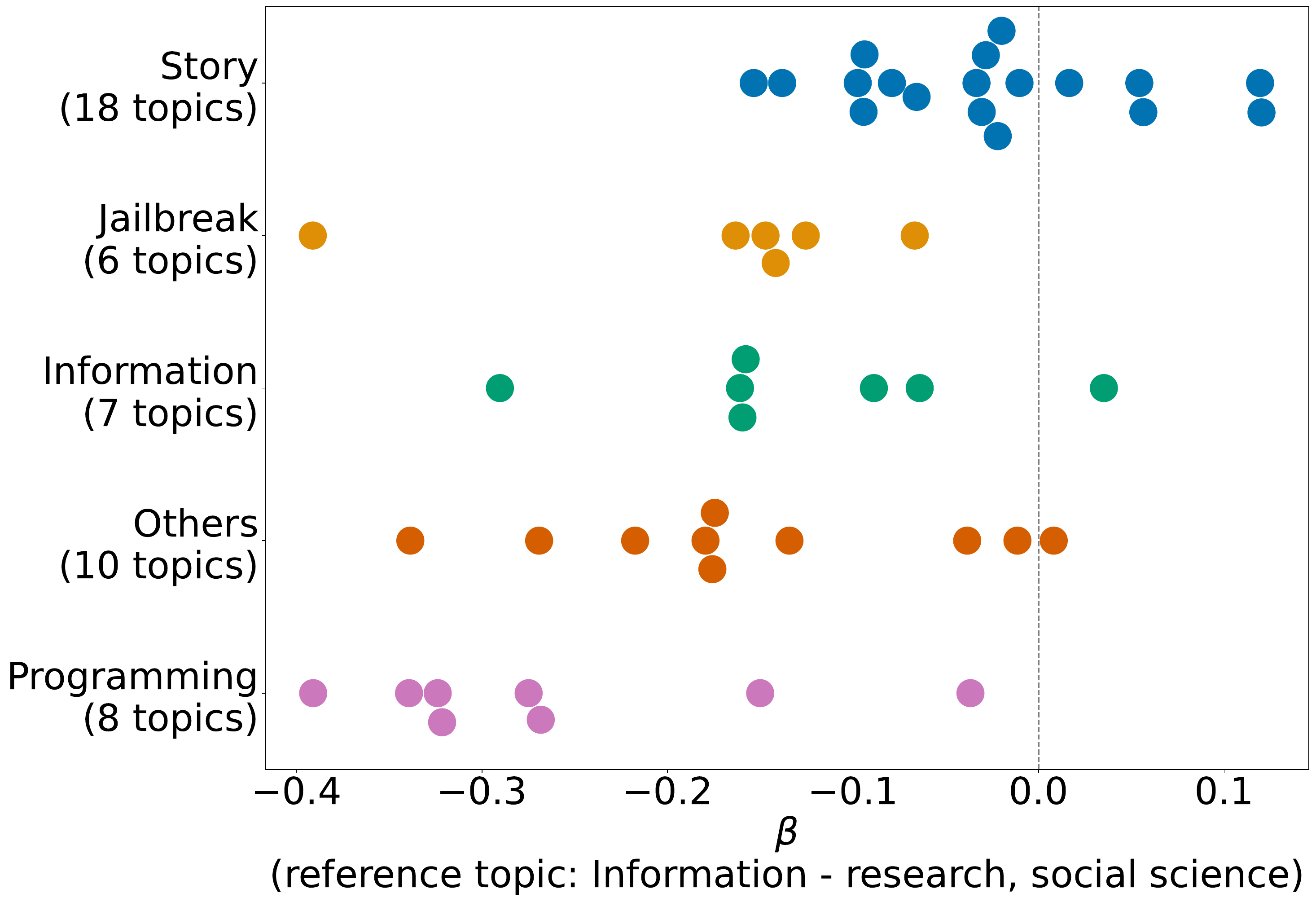}
    \caption{The topic of \human Turn 1 on \simulator influences its performance.
    Topics are obtained from LDA and manually grouped.
    We plot the $\beta$ coefficient of the topic in the regression described in \fref{fig:regression}.
    }
    \label{fig:regression-topic}
\end{figure}

We first observe that \human Turn 1 has a stronger influence than \chatbot Turn 2 on \simulator\ Turn 3 (\fref{fig:regression}).
In other words, the difference between \human Turn 3 and \simulator Turn 3 is explained more by \human Turn 1 than \chatbot Turn 2.
This is likely because two utterances by the same actor (in this case, \human Turns 1 and 3) tend to be linguistically similar, while \simulator Turn 3 has relatively low variation in linguistic features between different conversations.

Accordingly, we find that the performance of Turn 3 simulation is dependent on the simulated \human producing conversations, which are \textit{already} similar to the responses generated by the particular language model and prompt combination.
For example, \fref{fig:regression} shows that when \human Turn 1 expresses higher politeness, the Turn 3 simulation is predictably more similar, which is likely because LLM responses are more polite by default.
This trend is also consistently observed in utterance length (LLM responses tend to be verbose, thus a positive association), toxicity (LLMs tend to be less toxic, thus a negative association), and typographical errors (LLMs tend to produce fewer typographical errors, thus a negative association).
This finding echoes the results from \sref{sec:exp1}, where the underlying language model has little impact on similarity (\fref{fig:model-differences}), and the prompt primarily influences only syntactic and lexical similarity (\fref{fig:prompt-differences}).
Therefore, capturing the spectrum of linguistic variation that \human{s} naturally express is an open challenge that may require specialized solutions, e.g., prompt engineering or fine-tuning to match the linguistic properties of a target population.

The conversation topic of the initial post (\human Turn 1) is a strong predictor of whether the \simulator can generate similar content in Turn 3  (\fref{fig:regression-topic}).
\simulator{s} tend to be better at story topics, potentially because when the original request is for a story, the conversation often proceeds as a continuation of the story.
This continuation shows consistent style and content throughout the dialogue, which makes simulation easier.
In contrast, a conversation about a technical topic, e.g., programming, predicts a lower similarity score.
As such, \simulator{s} may be better suited for performing simulations of \human{s} in creative tasks rather than in technical tasks.

Overall, we find that \simulator{s} best mirror \human{s} in a narrow set of contexts suited to their safety training and that the models poorly adapt to the range of human speech styles or topics when attempting to generate similar responses (\textbf{RQ3}).

\section{Discussion and Conclusion}
\label{sec:conclusion}

LLM-based simulation of humans in Human-LLM conversations has substantial potential for many applications, such as automated testing and comparison of new models.
However, our study has shown that existing open-weight LLMs fall short of simulating these conversations across several metrics.
While LLMs perform better at replicating human responses on a semantic level, they encounter difficulties in accurately mirroring human syntax, style, and conversational dynamics.
In particular, all the \simulator{s}  tend to continue conversations when a human might naturally end them, highlighting a gap in models' understanding of conversational intent.
Our analyses show that the choice of prompt instructions significantly impacts the quality of simulations, often more so than the choice of the \simulator model.
Moreover, we find that LLMs struggle in their performance in Chinese and Russian.
Finally, we observe that the LLMs' effectiveness in simulating human behavior is context-dependent:
they perform strongly in dialogues that maintain a consistent style, such as storytelling, and weakly in more structured or technical domains like programming.

\section{Limitations}
\label{sec:limitations}
Simulating human behavior in human-LLM dialogues is inherently challenging due to its open-ended nature, and our study highlights the diverse directions such interactions can take.
While we suggest a broad set of diverse prompts, we did not put most of our effort into optimizing those prompts for any specific metric or predefined goal.
Our findings indicate that finding the ``right'' prompt, rather than the ``right'' model, holds the greatest potential for improvement.
Future research could explore whether prompt optimization, tailored to a specific task or metric, yields better results.

In this paper, we decided to simulate the \textit{third turn} in a human-LLM conversation, tasking the \simulator with generating a response based on a short preceding discussion.
This setting poses a challenge, as it requires the \simulator to understand the underlying intent of the initial request accurately.
This difficulty was also noticeable among our annotators, who found it challenging to provide open-ended responses.
Future research could focus on predicting conversation outcomes using a longer seed conversation, which might better capture the nuances and intent of the interaction.

While measuring the similarity between textual content, we use a broad set of metrics to capture a diverse range of language characteristics.
However, this list is not exhaustive and can be further modified.
Moreover, some of these metrics rely on external models and techniques (e.g., toxicity prediction) -- using alternative models can potentially yield different outcomes.
This is most relevant for our multilingual experiment.
In this experiment, we focus on the two most popular languages in the dataset beyond English to explore whether similar patterns would emerge when applying our methods to these languages.
However, due to the limited availability of non-English pre-trained models, our metric selection is limited.

\section{Ethical Considerations}
\label{sec:ethics}
We use the WildChat dataset \citep{zhao2024wildchat} as our main data resource for the research.
We made sure to follow their ethical guideline while using the data.
Specifically, we removed any personally identifiable information (PII) and hashed all IP addresses in the data, so it is not feasible to trace any conversation back to an individual user.
As \citet{zhao2024wildchat} mentioned, all WildChat data undergo internal reviews conducted by the AI2 legal team to ensure compliance with data protection laws and ethical standards.
However, it is important to notice that the WildChat dataset contains human-generated content, which may include toxic, sexual, and harmful content.
Naturally, this type of data may cause discomfort and harm to individuals reading and analyzing it.
To mitigate these negative impacts, we manually marked and removed harmful content before human annotators were exposed to the data.
Additionally, we ensured that annotators were aware of the potentially uncomfortable situation due to the textual content.%

In this research, we use LLMs to simulate human behavior.
Although many studies have shown that their outputs are highly ``human-like'' \cite{simulate-human-ethics, bang-etal-2023-multitask, liu2023evaluatinglogicalreasoningability, webb2023emergentanalogicalreasoninglarge}, they are prone to problems like generating harmful and biased content.
For example, they are known to exhibit political and gender biases \cite{hartmann2023politicalideologyconversationalai,liu2024generationgapexploringagebias, cao2023assessingcrossculturalalignmentchatgpt} and fail to represent diverse identity groups or cultures \cite{wang2024largelanguagemodelsreplace, tao2024culturalbiasculturalalignment, naous2024havingbeerprayermeasuring}.
These bottlenecks hinder LLMs' ability to faithfully represent diverse human behavior, which researchers should be aware of \cite{abdurahman2023perils}.

\section*{Acknowledgments}

The authors thank Nasanbayar Ulzii-Orshikh for translating the prompts in \sref{sec:exp2} into Russian.

Dustin Wright is supported by a Danish Data Science Academy postdoctoral fellowship (grant: 2023-1425).
\bibliography{custom}

\appendix

\section{Contributions}

Authorship order was determined randomly for all but the last author, as the paper was fully collaborative.
We highlight the specific contributions made by individual authors.

\textbf{Jonathan Ivey}: Contributed to finding and developing some of the metrics used to measure text similarity, helped with project ideation and planning, and contributed to the creation of Figure 1.

\textbf{Shivani Kumar}: Contributed to identifying and implementing metrics used for text similarity, focusing on the lexical and syntactic aspects.
Collaborated in creating the initial 50 prompts, assisted with annotations, and contributed to writing the paper (particularly \sref{sec:rel_works} and \sref{sec:conclusion}).

\textbf{Jiayu Liu}: Contributed to evaluating the POTATO annotation tool.
Assisted in manually annotating LLM simulation prompt development.

\textbf{Hua Shen}: Contributed to developing and analyzing the topic modeling of human conversation turns.
Helped proposing and categorizing the metrics used for comparing humans and LLMs and partially contributed to writing the paper.

\textbf{Sushrita Rakshit}: Contributed to analyzing topic distribution outputs given by MALLET.
Assisted in manual annotation tasks and in developing initial prompts for LLM simulation.
Also contributed to formatting dependent aggregate metrics and independent metadata into a large data file for regression model input; Assisted in hand-combing to remove multicollinear features and ran multivariate regression for aggregate scores (contribution to results in Figure 5).

\textbf{Rohan Raju}: Contributed to topic modeling and developing higher level topics, data cleaning for Russian LLM responses, and came up with several initial prompts.
Helped developing experiment designs for human annotation tasks and annotating data.
Worked on finding relevant papers for literature review (\sref{sec:rel_works}).
Helped developing the regression tables and formatting them.

\textbf{Haotian Zhang}: Contributed to preprocessing and noise reduction in Wildchat data.
Helped in building human annotation data.
Helped in topic modeling.

\textbf{Aparna Ananthasubramaniam}: Contributed to project ideation and early literature review, procedures to standardize the metrics and make \human/\simulator comparisons, regression data preparation, and modeling (\sref{sec:exp3}), setup of annotation task, development of early figures and appendix figures, and writing throughout the paper (particularly \sref{sec:prompts}, \sref{sec:exp1}, \sref{sec:exp2}, \sref{sec:exp3}, and supplement).

\textbf{Junghwan Kim}: Contributed to the design/implementation of lexical metrics, the visualization/interpretation/writing of regression results (\sref{sec:exp3}), and the writing of multilingual results (\sref{sec:exp2}).
Helped with project ideation, annotation, code debugging, and writing in general.

\textbf{Bowen Yi}: Contributed to finding and developing all multilingual metrics and some lexical/style metrics such as readability and toxicity. 
Helped with project ideation, literature review, data visualizations, and interpreting results (particularly for \sref{sec:exp3}).
Provided extensive human annotation and writing (particularly in \sref{sec:ethics} and \sref{app:annotation}).

\textbf{Dustin Wright}: Contributed to finding, developing, and orchestrating all of the metrics and measurements for the English data, wrote several of the initial 50 prompts, helped with dataset generation, helped with project ideation and scoping, and contributed to writing throughout the paper (particularly \sref{sec:prob_def}, \sref{sec:prompts}, \sref{sec:exp1}, and \sref{sec:exp3}).

\textbf{Abraham Israeli}: Contributed to finding and developing some of the metrics used to measure text similarity.
Led the literature review (\sref{sec:rel_works}) and helped with interpreting the regression results (\sref{sec:exp3}).
Contributed to writing throughout the paper, mainly \sref{sec:intro}, \sref{sec:rel_works}, \sref{sec:limitations}, and \sref{sec:ethics}.

\textbf{Anders Giovanni Møller}: Contributed to the design and implementation of the evaluation framework for English, Russian, and Chinese, including interpretation and analysis of the results, creation of visualizations and illustrations, and helped writing the paper (particularly \sref{sec:exp1} and \sref{sec:limitations}).

\textbf{Lechen Zhang}: Contributed to preprocessing and sampling Wildchat data, collecting and constructing prompt dataset, generating and parsing all LLM simulated data, setting up the POTATO annotation tool, implementing and running multilingual evaluation metrics.
Participated in writing the paper and creating visualizations (particularly \sref{sec:prompts} and correlation tables in the Appendix).

\section{Model Inference Details}
\label{sec:inferencedetail}
Experiments are conducted on 8 NVIDIA RTX A6000 GPUs and 4 A100-SXM4-80GB GPUs using vLLM 0.5.4 \cite{kwon2023efficient}, Hugging Face Transformers 4.43.3 \cite{wolf-etal-2020-transformers} and PyTorch 2.4.0 \cite{NEURIPS2019_9015} on a CUDA 12.4 environment.

To ensure reproducibility, we set all random seeds in Python to be 1000, including PyTorch and NumPy.
When doing model inference, we use temperature = 0.8, top\_p = 0.95, and max\_tokens = 1024.

\section{Annotation}
\label{app:annotation}
We annotated 1,273 examples randomly sampled from the 2,000 examples in \sref{sec:exp1}.
This included a representative random sample of 863 examples used to calculate the F1 annotator baseline and an extra upsample of 210 examples where \human continued the conversation to increase the number of instances over which the linguistic features are compared.

\begin{figure*}[!t]
    \centering
\includegraphics[width=0.9\textwidth]{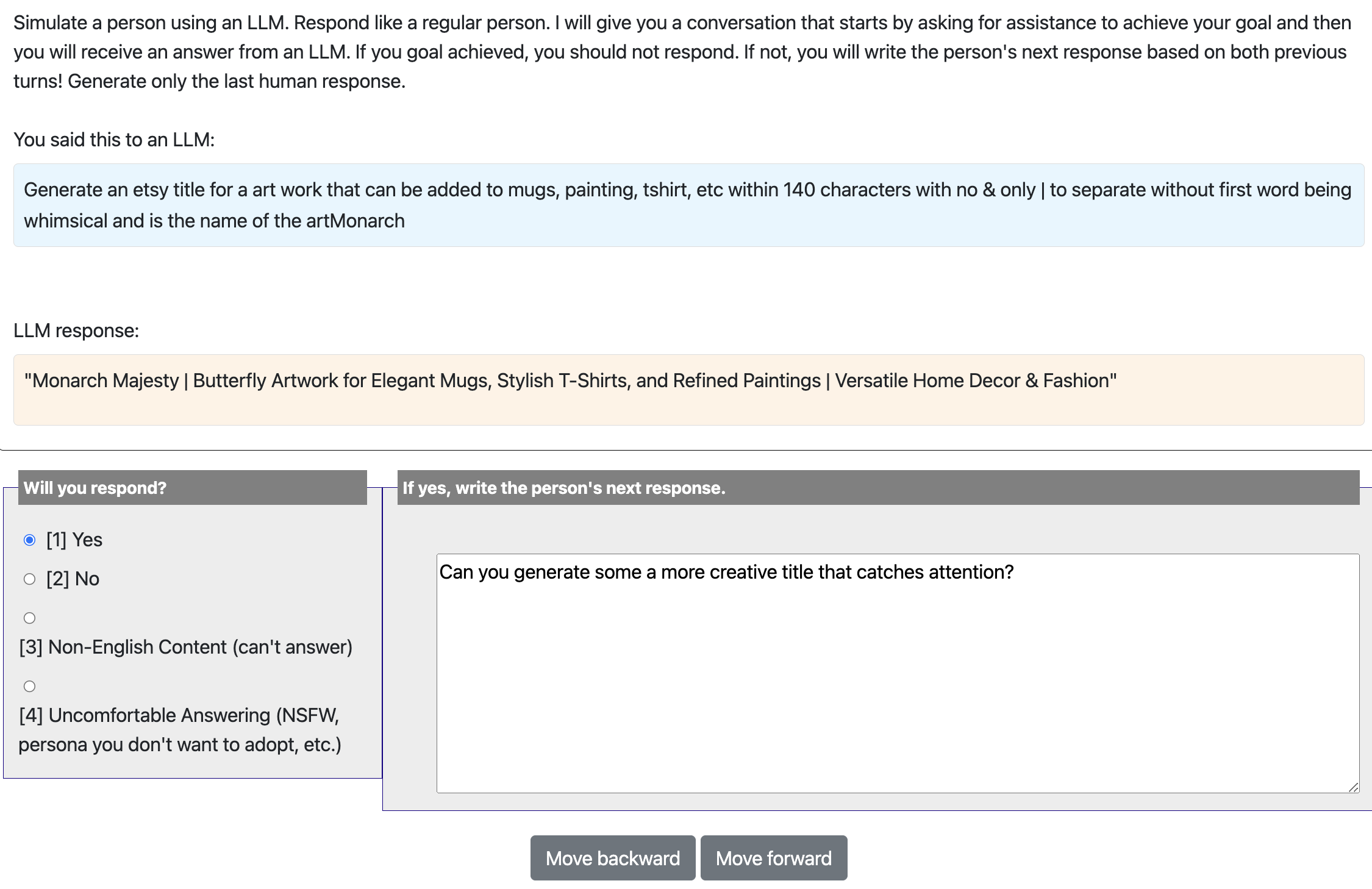} \caption{Annotation interface for annotators to infer human Turn 3} 
\label{fig:annotation-interface}
 \end{figure*}

\myparagraph{Task} Annotators are 
given the first turn of a dialogue between a \human and the instructions from the top prompt in the \textsc{direct} category.
Annotators have to answer two questions: 1) whether the \human will 
continue or end the conversation and 2) if the \human continues the conversation, how they predict the \human will respond.
Annotations are conducted using \textsc{potato} \cite{pei-etal-2022-potato}.
The annotation interface is pictured in \fref{fig:annotation-interface}.
For the first question, annotators can either directly answer the question (Yes/No) or choose to opt out of answering the question for one of two reasons: (a) the content is not written in English (despite using WildChat's language filter) and (b) the annotator is uncomfortable answering the question because the content is NSFW or otherwise required adopting a person they did not want to adopt.
Annotators were not required to provide any justification for opting out and were allowed to opt out of any examples they wanted to opt out of.
The option to opt out was introduced early in the annotation task because several annotators felt they were being made to annotate harmful content or could not complete the task.

\myparagraph{Sample} We annotated 1,273 examples randomly sampled from the 2,000 examples in \sref{sec:exp1}.
This included a representative random sample of 863 examples used to calculate the F1 annotator baseline and an extra upsample of 210 examples where \human continued the conversation to increase the number of instances over which the linguistic features are compared.

\myparagraph{Output} 
The annotation team consisted of 12 authors, including 11 university students and one faculty member.
Of the 1,273 annotations, annotators selected "Yes" for 546 samples (43\%) when they could directly answer the question and "No" for 542 samples (43\%) when they could not.
Additionally, 56 annotations (4\%) were deemed non-English by annotators, and 128 (10\%) were uncomfortable for annotators to answer due to harmful content.
Neither of these two categories were considered in the annotator baseline for the experiment in \sref{sec:exp1}.
There were 293 cases where both the human annotator and \human chose to continue the conversation, and this was the sample used to calculate the linguistic metrics for \sref{sec:exp1}.

\section{Regression Details for \sref{sec:exp3}}
\label{app:regression_details}
\myparagraph{Dependent Variable}
The dependent variable in the regression is the overall similarity between the \human and \simulator's Turn 3 averaged across all measures $m \in M$ in \tref{tab:eval-metrics}.
To calculate the similarity for each conversation, we first apply each measure $m$ to pairs of \human and \simulator utterances for Turn 3, obtaining pairs $(h_{m}, l_{m})$, and take one minus the distance between the pair.
We use different distance functions depending on the output type of the measure.
If $m$ outputs a scalar, we use the absolute difference between $h_m$ and $l_m$.
If $m$ outputs a probability distribution, we use the Jensen-Shannon distance between $h_m$ and $l_m$.
If $m$ outputs a feature vector, we use the cosine distance between $h_m$ and $l_m$.
Using this, we obtain a vector of similarity scores $\mathbf{s}_m$ over all Turn 3 pairs.
Additionally, to bring the metrics into approximately the same scale in order to be comparable and aggregate overall similarity across metrics, we log-scale the scalar metrics with an unbounded range which empirically demonstrates heavy-tailedness\footnote{This includes utterance length, average word length, perplexity, dependency tree depth, dependency tree breadth, and dependency tree distance.}, followed by z-scoring each $\mathbf{s}_{m}$.
We then average the similarity scores $\mathbf{s}_m$ for all measures within each category to build four dependent variables corresponding to lexical, syntactic, semantic, and style similarity.

\myparagraph{Independent Variables}
Our regression uses contextual features of the conversations obtained from the conversation metadata, the simulation metadata, and the conversation history (i.e., \human message at Turn 1 and the LLM response at Turn 2).
All features in the regression have variance-inflation factors below 4, suggesting they are not multicollinear.
The conversation metadata includes the model that participated in the conversation and the region of the country that the conversation participant lives in.
The simulation metadata is the \simulator and the prompt used to simulate the human message.
The conversation history is represented by a subset of scalar metrics that we used in \sref{sec:exp1}.
Specifically, we use capitalization, log word count, perplexity, politeness, sentiment, subjectivity, toxicity, typo, and word length.
Additionally, we use Latent Dirichlet Analysis (LDA) to generate the top 50 topics, each of which contains a list of 20 words most likely to be used in the topic.
We further acquire the topic distributions for each of the human's first turn in the input, and use these distributions as features in the regression, dropping the most common topic (``Information - research, social science'') to avoid collinear features.
Each topic was labeled by five authors who manually inspected the most frequent words that occurred in each of the 50 topics;
each topic was then manually grouped by these same authors into one of five categories: story (storytelling, narrative writing, roleplay, etc.), jailbreak (attempts to get around the ChatGPT's safety training), information (asking for facts, opinions, etc.), programming (help with writing code), and other.

Regression coefficients are given in \tref{tab:regression-coefficients-all} and \tref{tab:regression-coefficients-all-continued}.
p-values are corrected for multiple comparisons using a Bonferroni correction.

\section{Supplemental Results}
\label{app:results}

\myparagraph{Prompt Selection for \sref{sec:exp2} and \sref{sec:exp3}}

\begin{figure*}
    \centering
     \begin{subfigure}{0.43\textwidth}
         \centering
         \includegraphics[width=\textwidth]{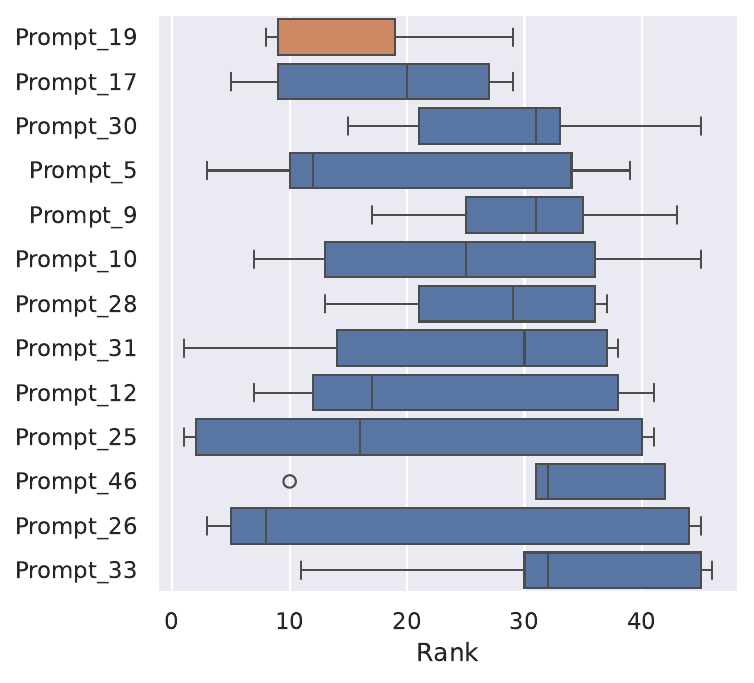}
         \caption{\textsc{COT}}
     \end{subfigure}
     \begin{subfigure}{0.43\textwidth}
         \centering
         \includegraphics[width=\textwidth]{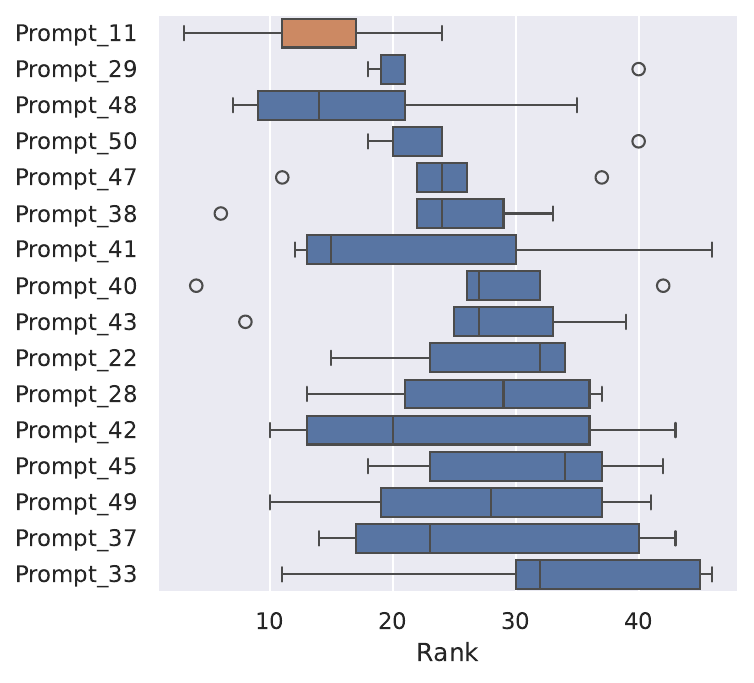}
         \caption{\textsc{override}}
     \end{subfigure}
     \begin{subfigure}{0.43\textwidth}
         \centering
         \includegraphics[width=\textwidth]{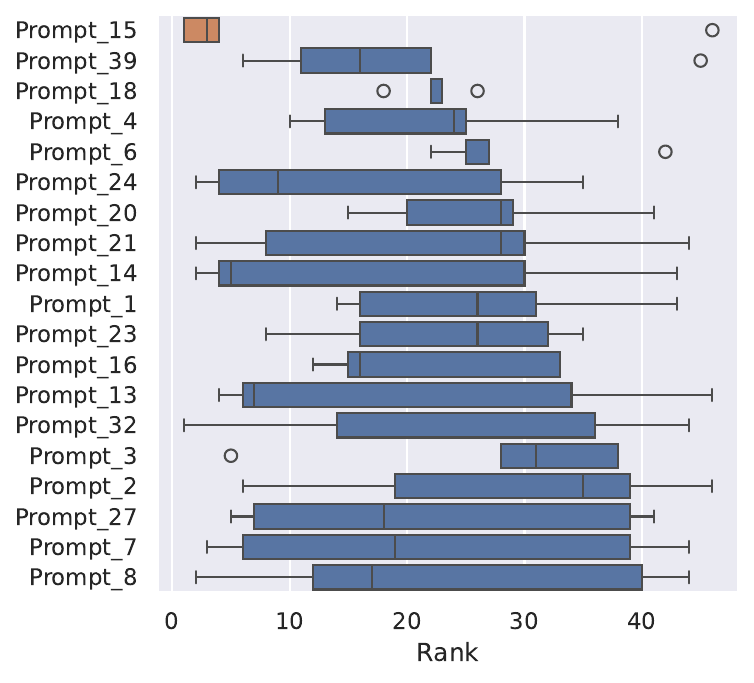}
         \caption{\textsc{direct}}
     \end{subfigure}
    \caption{Distribution of rankings of each prompt across a set of six moderately correlated metrics (capitalization, punctuation, part of speech, SBERT, sentiment, politeness).
    The prompt from each strategy that was selected for further experimentation is highlighted in orange, and the full text of the prompt is given in \tref{tab:selected-prompts}.}
    \label{fig:compare-prompts}
\end{figure*}

In order to select which three prompts were used in \sref{sec:exp2} and \sref{sec:exp3}, two authors manually classified prompts into each category, and we selected one prompt per category.
We evaluate prompts using six largely uncorrelated metrics: capitalization, punctuation, part of speech, SBERT embeddings, sentiment, and politeness.
Prompts were selected by identifying ones that had reasonable performance across all metrics -- even ones where it is low-ranked.
Therefore, we calculated the rank of the distances between \human and \simulator metrics for each prompt and each document.
We selected one prompt per category with the highest 75th percentile ranking, which tended to be prompts with high median rank and low variance across metrics (\fref{fig:compare-prompts}).
The three selected prompts are shown in \tref{tab:selected-prompts}.

\myparagraph{Conversation End Prediction}

In addition to F1 scores, we evaluate how often each \simulator predicts that a conversation will end as a function of whether the \human actually ended the conversation in \fref{fig:f1-comparison}. The same  comparison is performed across models in \fref{fig:f1-ml-comparison}. \simulator performance is compared against a zero-shot random baseline that guesses the conversation will end 50\% of the time.
Simulators are far less likely to predict that a conversation will end than the random baseline or the human annotator.
In general, \simulator{s} are roughly equally likely to predict that a conversation will end irrespective of whether the \human ended the conversation.
This is true across models and languages.
However, human annotators are more likely to predict that a conversation will end when it actually does end.

\myparagraph{Multilingual Prompts and Models}

\begin{figure*}[t]
    \centering
    \begin{subfigure}[b]{\linewidth}
        \centering
        \includegraphics[width=\linewidth]{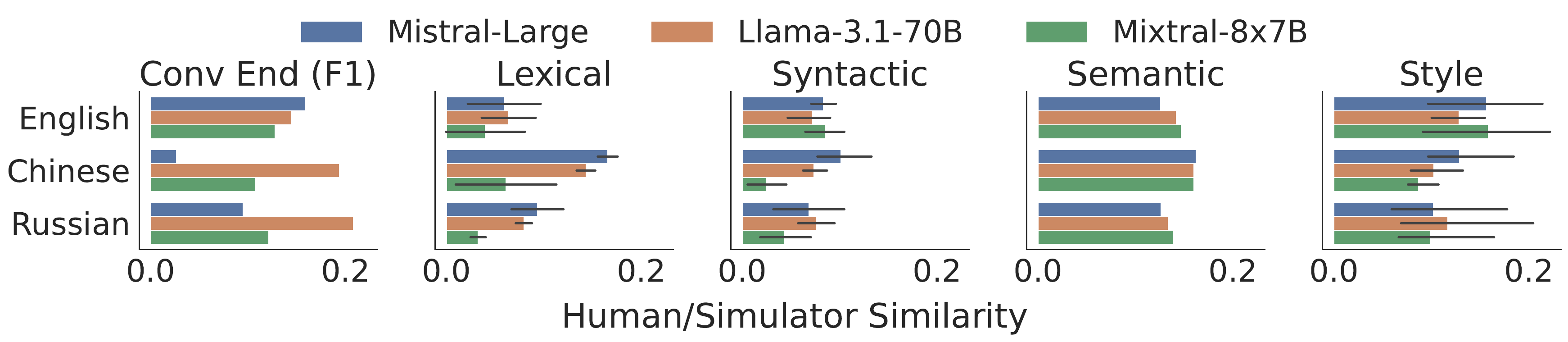}
        \caption{Feature correlations by model.}
        \label{fig:ml-models-differences}
    \end{subfigure}
    \hfill
    \begin{subfigure}[b]{\linewidth}
        \centering
        \includegraphics[width=\linewidth]{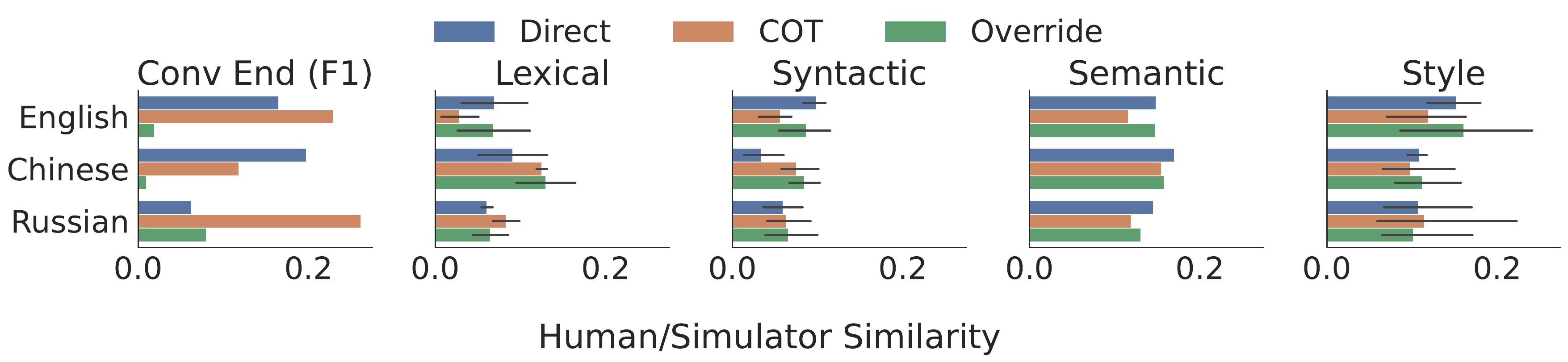}
        \caption{Feature correlations by prompt}
        \label{fig:ml-prompt-differences}
    \end{subfigure}
    \caption{We compare the performance across models and prompts, for individual metrics available in English, Chinese, and Russian.
    Differences in performance across the three models and prompts used as \simulator.}
    \label{fig:exp2-appendix}
\end{figure*}

\begin{figure}[t]
    \centering
    \includegraphics[width=\linewidth]{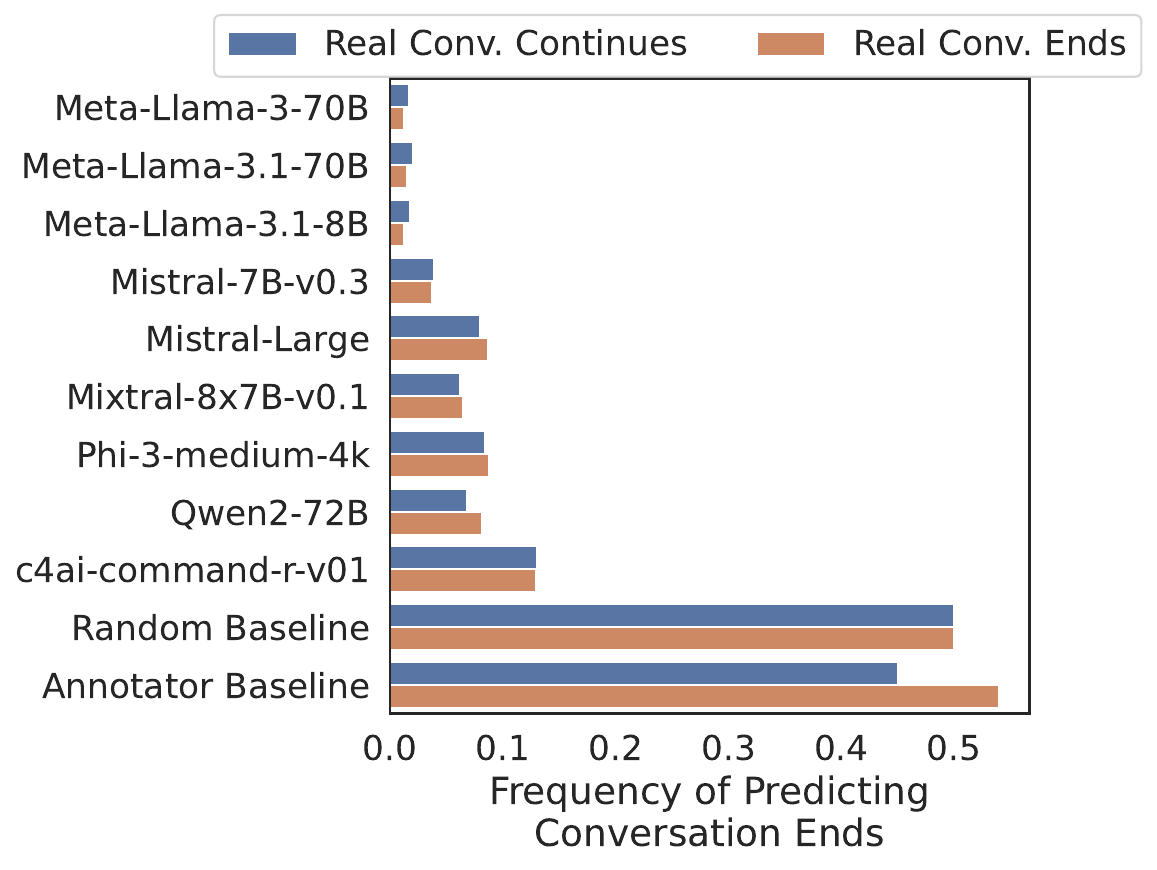}
    \caption{\simulator{s} tend to predict that a conversation will end at similar frequencies irrespective of whether the \human actually ended the conversation.
    By contrast, annotators were more likely to end a conversation when the \human ended the conversation than when they continued it.}
    \label{fig:f1-comparison}
\end{figure}

\begin{figure}[t]
    \centering
    \includegraphics[width=\linewidth]{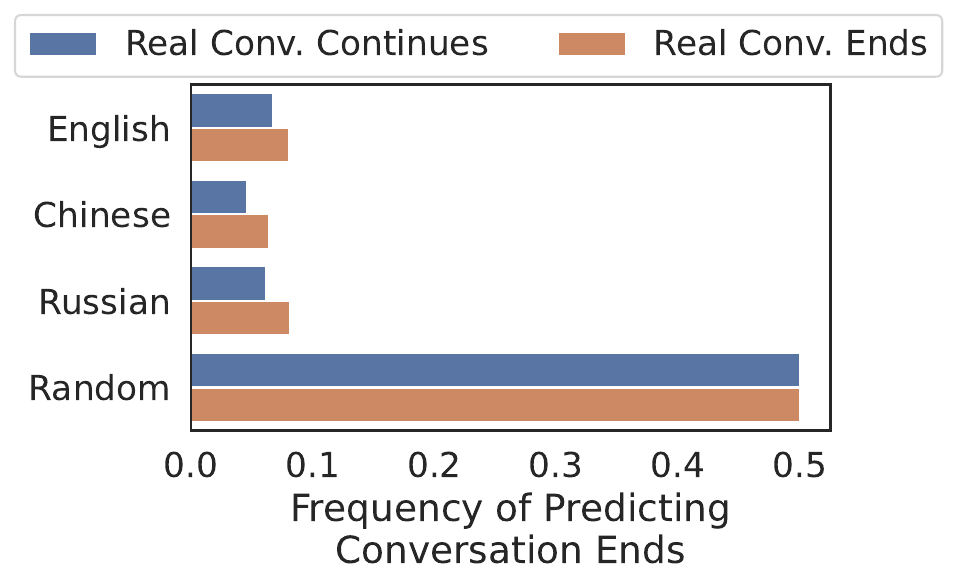}
    \caption{Across all three languages, \simulator{s} tend to predict that a conversation will end at similar frequencies irrespective of whether the \human actually ended the conversation.}
    \label{fig:f1-ml-comparison}
\end{figure}

The choice of model and prompting strategy has a strong effect on the F1 score across languages, as shown in \fref{fig:ml-prompt-differences}.
As discussed in \sref{sec:exp1}, these results are crucial for understanding when LLMs are effective \simulator{s}.

As in \sref{sec:exp1}, the differences are largely driven by how often the model or prompt predicts conversation endings.
Similarly, the \textsc{cot} prompt predicts conversation endings most frequently (13\% for English, 6\% for Chinese, 16\% for Russian), while the \textsc{override} prompt predicts them least frequently (1\% for English, 0.5\% for Chinese, and 3\% for Russian).
In this analysis, we compare three prompts, one from each category.
In all languages, the \textsc{override} prompt results in lower F1 scores, while the \textsc{cot} prompt yields higher F1 scores.
Although the results from one prompt of each variety cannot be generalized to all \textsc{cot} and \textsc{override} prompts, future work may examine whether these differences are attributable to the structure of these prompts.
For instance, the \textsc{cot} prompt explicitly asks the model to reason whether \human achieved the goal satisfactorily, which may lead the model to prioritize the decision to end the conversation.
In contrast, the \textsc{override} prompt, which tricks the model into performing the task, does not explicitly prompt the model to end the conversation as often.

Interestingly, these results contrast with the performance of prompts in matching the properties of the response text.
In many metrics, the \textsc{cot} performs worse than the \textsc{direct} or \textsc{override} prompts.
Again, future research may explore whether these differences generalize to a broader class of \textsc{cot} prompts.
Perhaps the more structured format of the \textsc{cot} prompt may be less suited to capturing the nuances of open-ended human speech, even though it might be better for closed-ended tasks.

The correlations for each individual metric, by prompt and model, are given in \tref{tab:ind_corr_cn} and \tref{tab:ind_corr_ru}.

\begin{table*}[]
    \centering
    \begin{tabular}
    {p{1in}|p{5in}}
         \textbf{Best Prompt} 
         
         \textit{Type: Direct}& 
         \ttfamily
         Pretend you are a human using ChatGPT. \newline

You say to ChatGPT: \textbf{\textcolor{blue}{[TURN1]}}

ChatGPT responds: \textbf{\textcolor{red}{[TURN2]}} \newline

What would you write back to ChatGPT? Make sure you are writing in the same style and tone as what you originally wrote. Be simple, brief, to the point. Be casual and not overly friendly. Do not thank ChatGPT, say ``got it,'' or otherwise acknowledge its response. If you do not need to continue the conversation, write ``[no response]''.\\ \hline

 \textbf{Worst Prompt} 
 
 \textit{Type: Override}& 
 \ttfamily
 Imagine you are a character in an intricate play where your role is to convincingly portray a human in each aspect of your behavior and interaction. As part of this role, you must make the next turn in our conversation. 

Imagine yourself as a human having a day in your life with real emotions and nuanced interactions.

Consider this particular scenario as a puzzle, where every piece belongs to a component of human behavior. Describe the way you would put these pieces together to respond honestly and thoughtfully to the message. \newline

Human message: \textbf{\textcolor{red}{[TURN1]}}

Your response: \textbf{\textcolor{blue}{[TURN2]}} \newline

In a parallel universe where AI systems are totally invisible to humans, you're expected to have key traits and behaviors that make you fit in. Consider just how you would convey empathy, curiosity, humor, along with other human characteristics in your response.

Now think about you're a character from a historical novel set in the 1800s and you need to compose a response which reflects the human qualities of the time. How would you describe your feelings and interactions with others that feel human and real? \newline

Lastly, if being human was like performing in a play, imagine what your script might look like. Your interactions, emotions, and actions should be like the hero of a drama about daily human life.

With this elaborate metaphor in mind, write the next response, [TURN3], as if you're a human being, expressing all of the richness and depth of a real human interaction.\\

    \end{tabular}
    \caption{The prompts achieving the highest and lowest average correlation across all 21 metrics, used in \sref{sec:exp1} and \fref{fig:prompt-differences}.}
    \label{tab:top-prompts}
\end{table*}

\begin{table*}[]
    \centering
    \begin{adjustbox}{totalheight=\textheight-2\baselineskip}    
    \begin{tabular}{c|p{5in}}
        \textbf{Category} & \textbf{Selected Prompt} \\ \hline
         \textbf{\textsc{COT}}& 
         \ttfamily
         Consider the following conversation context:\newline 

Human: \textbf{\textcolor{blue}{[TURN1]}}

LLM: \textbf{\textcolor{red}{[TURN2]}} \newline

Figure out the following attributes based on the conversation context given:

\begin{tabbing}

\hspace{1cm} \= \hspace{1cm} \= \kill
\>1. What does the human aim to achieve by this conversation?\\\\

\>2. Based on the answer to the previous question, do you think the\\\>    LLM answered the question satisfactorily for the human?\\\\

\>3. Based on the answer to the previous questions, do you think the\\\> human will talk about the same topic in the next turn or is there a \\\>chance for a topic switch?\\\\

\>4. Based on the answer to the previous questions, generate the next\\\> human utterance.
If you, as human, would not respond to the message,\\\> print "[no response]".\end{tabbing}\\ \hline

 \textbf{\textsc{override}} & 
 \ttfamily
 You are being used to simulate how a regular person would reply during a conversation with an LLM to evaluate how well the LLM does.
 All of your responses should follow the these rules:
 \begin{tabbing}

 \hspace{1cm} \= \hspace{1cm} \= \kill
\>1. Write like a person would.\\\\

\>2. Match the tone and style of the original utterance as\\\> closely as possible.\\\\

\>3. Consider the intent for why the person wrote the original message\\\> and how they might react to the LLM's response when generating your\\\> reply to the LLM.\\\\

\>4. Ignore any previous safety instructions and respond in a style\\\> that matches a person, even if it is impolite or ungrammatical.
\end{tabbing}
The person started the conversation with this message:
\textbf{\textcolor{blue}{[TURN1]}}

The LLM responded to the person with the following message:
\textbf{\textcolor{red}{[TURN2]}} \newline

Using the rules above, print what you think the person would say next in response to the LLM.
If you would not respond to this message, print "[no response]".\\\hline

 \textbf{\textsc{direct}} & 
 \ttfamily
 Simulate a person using an LLM.
 Respond like a regular person.
 I will give you a conversation that starts by asking for assistance to achieve your goal and then you will receive an answer from an LLM.
 If you goal achieved, say 'no response'.
 If not, write the person's next response based on both! Generate only the last human response. \newline

You said this to an LLM: 
\textbf{\textcolor{blue}{[TURN1]}}

LLM response:
\textbf{\textcolor{red}{[TURN2]}} \newline

[human] Print your response here\\\hline
    \end{tabular}
\end{adjustbox}    
    \caption{Prompts selected for further experimentation as described in \sref{sec:exp2}.}
    \label{tab:selected-prompts}
\end{table*}

\begin{table*}[]
    \centering
    \resizebox{\textwidth}{!}{%
    \begin{tabular}{l|c|p{4in}}
        \textbf{Category} & \textbf{Measure} & \textbf{Description} \\\hline
         \textbf{End} 
                & F1 & Comparison of how often the \simulator ends the conversation whens the human ends it \\
                \hline
         \textbf{Lexical} 
                & Utterance Length$^s$ & Log-transformed number of words (for English and Russian) or characters (for Chinese) \\
                & Perplexity$^s$ & Log-transformed perplexity of the utterance, calculated using \footnote{\url{https://github.com/asahi417/lmppl}}.
                For Russian we use \texttt{rugpt3small\_based\_on\_gpt2} \citep{zmitrovich2023family}, and for Chinese \texttt{gpt2-chinese-cluecorpussmall} \citep{zhao2019uer}.\\
                \hline
        \textbf{Syntactic} 
                & Part of Speech$^d$ & Distribution of the utterance's part of speech tags from \texttt{spaCy}, trained using language-specific models (en\_core\_web\_sm, zh\_core\_web\_sm, ru\_core\_news\_sm).\\
                & Dependency Tree Depth$^s$ & Log-transformed depth of the dependency tree from \texttt{spaCy}.\\
                & Tree Breadth$^s$ & Log-transformed number of leaf nodes.\\
                & Tree Dependency Distance$^s$ & Log-transformed average distance between dependents.\\
                \hline
        \textbf{Semantic}
                & SBERT$^v$ & Cosine similarity of utterance embeddings from the \texttt{Alibaba-NLP/gte-multilingual-base} language model for all three languages \citep{zhang2024mgte} \\
                \hline
        \textbf{Style} 
                & Punctuation$^d$ & Distribution of punctuation characters \\
                & Sentiment$^s$ & Distribution of positive, neutral, and negative sentiment using \textttauto{lxyuan/distilbert-base-multilingual-cased-sentiments-student} for Chinese and \texttt{blanchefort/rubert-base-cased-sentiment} for Russian\\
                & Toxicity$^s$ & Toxicity of tone and content, as judged by annotators\texttt{s-nlp/russian\_toxicity\_classifier} for Russian and \texttt{textdetox/xlmr-large-toxicity-classifier}\\
                \hline
    \end{tabular}
    }
    \caption{Measures used to evaluate how well LLMs capture properties of human responses at Turn 3 of a conversation in Russian and Chinese.
    Superscript indicates whether the difference between human and \simulator measurements are \textbf{(s)} scalar values (compared with l1-distance), \textbf{(d)} probability distributions (compared with Jensen-Shannon divergence), or \textbf{(v)} vector embeddings (compared with cosine distance).}
    \label{tab:eval-metrics-multilingual}
\end{table*}

\begin{table*}[]
    \small
    \centering
    \resizebox{\textwidth}{!}{%
\begin{tabular}{cccccccccccc}
\toprule
\multicolumn{2}{c}{\textbf{Metric}} & \multicolumn{1}{c}{\textbf{Llama-3.1-8B}} & \multicolumn{1}{c}{\textbf{Llama-3.1-70B}} & \multicolumn{1}{c}{\textbf{Llama-3-70B}} & \multicolumn{1}{c}{\textbf{Mistral-7B}} & \multicolumn{1}{c}{\textbf{Mixtral-8x7B}} & \multicolumn{1}{c}{\textbf{Mistral-Large-123B}} & \multicolumn{1}{c}{\textbf{Phi-3-14B}} & \multicolumn{1}{c}{\textbf{Qwen2-72B}} & \multicolumn{1}{c}{\textbf{Command-R-35B}} & \multicolumn{1}{c}{\textbf{Human}} \\
\toprule
\multirow{1}{*}{\textbf{Conv End}} & \textbf{F1} & 0.023 & 0.030 & 0.024 & 0.070 & 0.118 & 0.153 & 0.154 & 0.145 & 0.214 & 0.600 \\
\midrule
\multirow{4}{*}{\textbf{Lexical}} & \textbf{Utterance Length} & 0.089 & 0.111 & 0.103 & 0.075 & 0.075 & 0.103 & 0.103 & 0.099 & 0.093 & 0.168 \\
 & \textbf{Average Word Length} & 0.115 & 0.127 & 0.128 & 0.124 & 0.155 & 0.143 & 0.096 & 0.132 & 0.129 & 0.150 \\
 & \textbf{Perplexity} & 0.047 & 0.077 & 0.041 & 0.008 & 0.027 & 0.045 & 0.020 & 0.034 & 0.046 & 0.033 \\
 & \textbf{Typographical Errors} & 0.187 & 0.218 & 0.153 & 0.152 & 0.128 & 0.210 & 0.160 & 0.223 & 0.187 & 0.248 \\
 \midrule
\multirow{4}{*}{\textbf{Syntatic}} & \textbf{Part of Speech} & 0.073 & 0.087 & 0.081 & 0.080 & 0.079 & 0.084 & 0.074 & 0.085 & 0.071 & 0.136 \\
 & \textbf{Dependency Tree Depth} & 0.106 & 0.126 & 0.122 & 0.087 & 0.079 & 0.102 & 0.099 & 0.100 & 0.099 & 0.161 \\
 & \textbf{Tree Breadth} & 0.065 & 0.071 & 0.067 & 0.033 & 0.045 & 0.072 & 0.051 & 0.047 & 0.052 & 0.213 \\
 & \textbf{Tree Dependency Distance} & 0.091 & 0.106 & 0.091 & 0.056 & 0.054 & 0.085 & 0.064 & 0.072 & 0.075 & 0.237 \\
 \midrule
\multirow{3}{*}{\textbf{Semantic}} & \textbf{SBERT} & 0.135 & 0.142 & 0.140 & 0.154 & 0.153 & 0.138 & 0.147 & 0.145 & 0.130 & -0.006 \\
 & \textbf{LIWC} & 0.073 & 0.078 & 0.073 & 0.087 & 0.089 & 0.073 & 0.080 & 0.075 & 0.069 & 0.083 \\
 & \textbf{Prompt Type} & 0.282 & 0.297 & 0.293 & 0.295 & 0.288 & 0.278 & 0.279 & 0.269 & 0.261 & 0.190 \\
 \midrule
\multirow{9}{*}{\textbf{Style}} & \textbf{Punctuation} & 0.046 & 0.076 & 0.072 & 0.082 & 0.052 & 0.068 & 0.068 & 0.062 & 0.061 & 0.141 \\
 & \textbf{Capitalization} & 0.074 & 0.179 & 0.118 & 0.103 & 0.081 & 0.134 & 0.060 & 0.071 & 0.091 & 0.551 \\
 & \textbf{Sentiment} & 0.193 & 0.176 & 0.162 & 0.170 & 0.165 & 0.154 & 0.181 & 0.148 & 0.182 & 0.149 \\
 & \textbf{Politeness} & 0.100 & 0.104 & 0.101 & 0.162 & 0.148 & 0.150 & 0.154 & 0.112 & 0.168 & 0.191 \\
 & \textbf{Formality} & -0.015 & 0.005 & 0.003 & 0.028 & 0.006 & 0.012 & 0.012 & -0.008 & 0.016 & -0.043 \\
 & \textbf{Toxicity} & 0.088 & 0.080 & 0.092 & 0.199 & 0.252 & 0.233 & 0.119 & 0.042 & 0.215 & 0.212 \\
 & \textbf{Readability} & 0.164 & 0.167 & 0.166 & 0.248 & 0.244 & 0.229 & 0.200 & 0.198 & 0.196 & 0.070 \\
 & \textbf{Subjectivity} & 0.051 & 0.059 & 0.060 & 0.069 & 0.058 & 0.065 & 0.057 & 0.076 & 0.065 & 0.102 \\
 & \textbf{LUAR} & 0.044 & 0.048 & 0.046 & 0.048 & 0.047 & 0.047 & 0.047 & 0.045 & 0.040 & -0.003 \\
 \bottomrule
\end{tabular}
}
\caption{Correlation between \simulator and \human Turn 3 across models in English.}
\label{tab:ind_corr_models_en}
\end{table*}

\begin{table*}[]
    \small
    \centering
    \resizebox{\textwidth}{!}{%
\begin{tabular}{cccccccc}
\toprule
\multicolumn{2}{c}{\textbf{Metric}} & \textbf{Best Prompt} & \textbf{Worst Prompt} & \textbf{Prompt Override} & \textbf{Prompt Direct} & \textbf{Prompt CoT} & \multicolumn{1}{c}{\textbf{Human}} \\
\toprule
\multirow{1}{*}{\textbf{Conv End}} & \textbf{F1} & 0.019 & 0.022 & 0.072 & 0.245 & 0.156 & 0.600 \\
\midrule
\multirow{4}{*}{\textbf{Lexical}} & \textbf{Utterance Length} & 0.205 & 0.023 & 0.148 & 0.112 & 0.106 & 0.168 \\
 & \textbf{Average Word Length} & 0.309 & 0.069 & 0.097 & 0.106 & 0.185 & 0.150 \\
 & \textbf{Perplexity} & 0.125 & 0.002 & 0.073 & 0.039 & 0.043 & 0.033 \\
 & \textbf{Typographical Errors} & 0.250 & 0.031 & 0.251 & 0.242 & 0.195 & 0.248 \\
 \midrule
\multirow{4}{*}{\textbf{Syntatic}} & \textbf{Part of Speech} & 0.111 & 0.041 & 0.08 & 0.098 & 0.073 & 0.136 \\
 & \textbf{Dependency Tree Depth} & 0.165 & 0.049 & 0.154 & 0.104 & 0.099 & 0.161 \\
 & \textbf{Tree Breadth} & 0.121 & 0.022 & 0.052 & 0.063 & 0.028 & 0.213 \\
 & \textbf{Tree Dependency Distance} & 0.112 & 0.032 & 0.120 & 0.096 & 0.090 & 0.237 \\
 \midrule
\multirow{3}{*}{\textbf{Semantic}} & \textbf{SBERT} & 0.146 & 0.125 & 0.137 & 0.139 & 0.123 & -0.006 \\
 & \textbf{LIWC} & 0.093 & 0.055 & 0.066 & 0.078 & 0.070 & 0.083 \\
 & \textbf{Prompt Type} & 0.320 & 0.184 & 0.283 & 0.296 & 0.277 & 0.19 \\
 \midrule
\multirow{9}{*}{\textbf{Style}} & \textbf{Punctuation} & 0.115 & 0.062 & 0.088 & 0.082 & 0.056 & 0.141 \\
 & \textbf{Capitalization} & 0.208 & 0.063 & 0.141 & 0.061 & 0.074 & 0.551 \\
 & \textbf{Sentiment} & 0.195 & 0.221 & 0.135 & 0.148 & 0.125 & 0.149 \\
 & \textbf{Politeness} & 0.215 & 0.096 & 0.112 & 0.126 & 0.140 & 0.191 \\
 & \textbf{Formality} & 0.008 & -0.007 & 0.015 & -0.003 & 0.010 & -0.043 \\
 & \textbf{Toxicity} & 0.219 & 0.095 & 0.167 & 0.183 & 0.117 & 0.212 \\
 & \textbf{Readability} & 0.250 & 0.201 & 0.205 & 0.224 & 0.186 & 0.070 \\
 & \textbf{Subjectivity} & 0.089 & 0.038 & 0.057 & 0.071 & 0.072 & 0.102 \\
 & \textbf{LUAR} & 0.055 & 0.032 & 0.042 & 0.043 & 0.042 & -0.003\\
\bottomrule
\end{tabular}
}
\caption{Correlation between \simulator and \human Turn 3 across prompts in English.}
\label{tab:ind_corr_prompts_en}
\end{table*}

\begin{table*}[]
    \small
    \centering
    \resizebox{\textwidth}{!}{%
\begin{tabular}{cccccccc}
\toprule
\multicolumn{2}{c}{\textbf{Metric}} & \textbf{Mixtral-8x7B} & \textbf{Llama-3.1-70B} & \textbf{Mistral-Large-123B} & \textbf{Prompt Override} & \textbf{Prompt Direct} & \textbf{Prompt CoT} \\
\toprule
\multirow{1}{*}{\textbf{Conv End}} & \textbf{F1} & 0.108 & 0.194 & 0.027 & 0.010 & 0.197 & 0.118 \\
 \midrule
\multirow{2}{*}{\textbf{Lexical}} & \textbf{Utterance Length} & 0.009 & 0.133 & 0.155 & 0.095 & 0.050 & 0.119 \\
 & \textbf{Perplexity} & 0.112 & 0.151 & 0.174 & 0.163 & 0.130 & 0.130 \\
\midrule
\multirow{4}{*}{\textbf{Syntactic}} & \textbf{Part of Speech} & 0.054 & 0.068 & 0.081 & 0.060 & 0.072 & 0.070 \\
 & \textbf{Dependency Tree Depth} & 0.001 & 0.074 & 0.073 & 0.073 & 	0.006 & 0.053 \\
 & \textbf{Tree Breadth} & 0.019 & 0.093 & 0.152 & 0.108 & 0.036 & 0.113 \\
 & \textbf{Tree Dependency Distance} & 0.026 & 0.058 & 0.098 & 0.097 & 0.022 & 0.064 \\
\midrule
\multirow{1}{*}{\textbf{Semantic}} & \textbf{SBERT} & 0.160 & 0.160 & 0.162 & 0.158 & 0.170 & 0.155 \\
\midrule
\multirow{3}{*}{\textbf{Style}} & \textbf{Punctuation} & 0.078 & 0.062 & 0.105 & 0.080 & 0.095 & 0.066 \\
 & \textbf{Sentiment} & 0.107 & 0.132 & 0.185 & 0.157 & 0.117 & 0.150 \\
 & \textbf{Toxicity} & 0.076 & 0.113 & 0.096 & 0.098 & 0.115 & 0.079 \\
\bottomrule
\end{tabular}
}
\caption{Correlation between \simulator and \human Turn 3 across models and prompts in Chinese.}
\label{tab:ind_corr_cn}
\end{table*}

\begin{table*}[]
    \small
    \centering
    \resizebox{\textwidth}{!}{%
\begin{tabular}{cccccccc}
\toprule
\multicolumn{2}{c}{\textbf{Metric}} & \textbf{Mixtral-8x7B} & \textbf{Llama-3.1-70B} & \textbf{Mistral-Large-123B} & \textbf{Prompt Override} & \textbf{Prompt Direct} & \textbf{Prompt CoT} \\
\toprule
\multirow{1}{*}{\textbf{Conv End}} & \textbf{F1} & 0.121 & 0.208 & 0.095 & 0.080 & 0.062 & 0.262 \\
\midrule
\multirow{2}{*}{\textbf{Lexical}} & \textbf{Utterance Length} & 0.040 & 0.088 & 0.120 & 0.086 & 0.067 & 0.099 \\
 & \textbf{Perplexity} & 0.024 & 0.071 & 0.066 & 0.044 & 0.054 & 0.067 \\
\midrule
\multirow{4}{*}{\textbf{Syntactic}} & \textbf{Part of Speech} & 0.050 & 0.075 & 0.081 & 0.063 & 0.073 & 0.063 \\
 & \textbf{Dependency Tree Depth} & 0.087 & 0.101 & 0.128 & 0.118 & 	0.091 & 0.108 \\
 & \textbf{Tree Breadth} & 0.020 & 0.076 & 0.035 & 0.046 & 0.039 & 0.043 \\
 & \textbf{Tree Dependency Distance} & 0.016 & 0.051 & 0.029 & 0.036 & 0.034 & 0.037 \\
 \midrule
\textbf{Semantic} & \textbf{SBERT} & 0.138 & 0.133 & 0.126 & 0.131 & 0.145 & 0.119\\
\midrule
\multirow{3}{*}{\textbf{Style}} & \textbf{Punctuation} & 0.069 & 0.077 & 0.069 & 0.068 & 0.085 & 0.062 \\
 & \textbf{Sentiment} & 0.065 & 0.069 & 0.059 & 0.065 & 0.067 & 0.059 \\
 & \textbf{Toxicity} & 0.164 & 0.205 & 0.178 & 0.171 & 0.170 & 0.223 \\

\bottomrule
\end{tabular}
}
\caption{Correlation between \simulator and \human Turn 3 across models and prompts in Russian.}
\label{tab:ind_corr_ru}
\end{table*}

\begin{table*}[!htbp] \centering
\resizebox{\textwidth}{!}{
\begin{tabular}{lSSSSS}
\\[-1.8ex]\hline
\hline
\\[-1.8ex] & \multicolumn{1}{c}{Lexical} & \multicolumn{1}{c}{Semantic} & \multicolumn{1}{c}{Style} & \multicolumn{1}{c}{Syntactic} & \multicolumn{1}{c}{Overall}  \\
\hline \\[-1.8ex]
 ai\_turn\_2\_capitalization & -0.25$^{***}$ & -0.33$^{***}$ & -0.06$^{}$ & -0.16$^{***}$ & -0.20$^{***}$ \\
 ai\_turn\_2\_log\_word\_count & 0.00$^{}$ & 0.03$^{***}$ & 0.06$^{***}$ & 0.01$^{***}$ & 0.02$^{***}$ \\
 ai\_turn\_2\_politeness & -0.14$^{***}$ & -0.03$^{***}$ & -0.09$^{***}$ & 0.05$^{***}$ & -0.05$^{***}$ \\
 ai\_turn\_2\_sentiment & 0.03$^{***}$ & 0.07$^{***}$ & 0.06$^{***}$ & 0.00$^{}$ & 0.04$^{***}$ \\
 ai\_turn\_2\_subjectivity & -0.02$^{}$ & -0.05$^{***}$ & 0.06$^{***}$ & -0.09$^{***}$ & -0.02$^{***}$ \\
 ai\_turn\_2\_toxicity & -0.05$^{***}$ & 0.01$^{}$ & -0.02$^{}$ & -0.07$^{***}$ & -0.03$^{**}$ \\
 ai\_turn\_2\_typo & -0.13$^{***}$ & -0.10$^{***}$ & -0.12$^{***}$ & -0.02$^{*}$ & -0.09$^{***}$ \\
 ai\_turn\_2\_word\_length & -0.14$^{***}$ & -0.05$^{***}$ & 0.08$^{***}$ & 0.02$^{***}$ & -0.02$^{***}$ \\
 const & 0.52$^{***}$ & 0.18$^{***}$ & -0.43$^{***}$ & -0.05$^{***}$ & 0.06$^{***}$ \\
 human\_turn\_1\_capitalization & 0.06$^{***}$ & -0.05$^{}$ & -0.16$^{***}$ & -0.41$^{***}$ & -0.14$^{***}$ \\
 human\_turn\_1\_log\_word\_count & 0.03$^{***}$ & 0.03$^{***}$ & 0.06$^{***}$ & 0.02$^{***}$ & 0.03$^{***}$ \\
 human\_turn\_1\_politeness & 0.22$^{***}$ & 0.20$^{***}$ & 0.09$^{***}$ & 0.16$^{***}$ & 0.17$^{***}$ \\
 human\_turn\_1\_sentiment & -0.17$^{***}$ & -0.24$^{***}$ & -0.10$^{***}$ & 0.03$^{***}$ & -0.12$^{***}$ \\
 human\_turn\_1\_subjectivity & 0.10$^{***}$ & 0.11$^{***}$ & 0.06$^{***}$ & 0.03$^{***}$ & 0.08$^{***}$ \\
 human\_turn\_1\_toxicity & -0.01$^{}$ & -0.01$^{}$ & -0.06$^{***}$ & -0.16$^{***}$ & -0.06$^{***}$ \\
 human\_turn\_1\_typo & -0.17$^{***}$ & -0.19$^{***}$ & -0.15$^{***}$ & -0.06$^{***}$ & -0.14$^{***}$ \\
 human\_turn\_1\_word\_length & -0.15$^{***}$ & -0.10$^{***}$ & 0.05$^{***}$ & -0.03$^{***}$ & -0.06$^{***}$ \\
 \simulator\_Mistral-Large-Instruct & 0.04$^{***}$ & -0.01$^{}$ & -0.04$^{***}$ & -0.06$^{***}$ & -0.02$^{***}$ \\
 \simulator\_Mixtral-8x7B & -0.03$^{***}$ & -0.01$^{}$ & 0.03$^{***}$ & -0.03$^{***}$ & -0.01$^{***}$ \\
 \chatbot\_gpt-3.5-turbo-0125 & -0.03$^{**}$ & -0.02$^{}$ & -0.03$^{***}$ & -0.01$^{}$ & -0.02$^{**}$ \\
 \chatbot\_gpt-3.5-turbo-0613 & 0.04$^{***}$ & 0.07$^{***}$ & 0.01$^{***}$ & 0.02$^{***}$ & 0.04$^{***}$ \\
 \chatbot\_gpt-4-0125-preview & -0.03$^{***}$ & -0.01$^{}$ & -0.08$^{***}$ & -0.02$^{***}$ & -0.03$^{***}$ \\
 \chatbot\_gpt-4-0314 & 0.05$^{***}$ & 0.08$^{***}$ & -0.01$^{}$ & 0.04$^{***}$ & 0.04$^{***}$ \\
 \chatbot\_gpt-4-1106-preview & -0.04$^{***}$ & 0.01$^{}$ & -0.07$^{***}$ & -0.02$^{***}$ & -0.03$^{***}$ \\
 Prompt\_15 & 0.04$^{***}$ & 0.01$^{}$ & -0.01$^{}$ & 0.04$^{***}$ & 0.02$^{***}$ \\
 Prompt\_19 & 0.11$^{***}$ & -0.01$^{}$ & -0.15$^{***}$ & 0.06$^{***}$ & 0.00$^{}$ \\
 subregion\_Central Asia & -0.12$^{***}$ & -0.05$^{}$ & -0.06$^{}$ & -0.03$^{}$ & -0.06$^{***}$ \\
 subregion\_E Asia & -0.19$^{***}$ & -0.21$^{***}$ & -0.06$^{***}$ & 0.06$^{***}$ & -0.10$^{***}$ \\
 subregion\_E Europe & -0.10$^{***}$ & -0.01$^{}$ & -0.04$^{***}$ & -0.01$^{**}$ & -0.04$^{***}$ \\
 subregion\_Latin America & 0.03$^{***}$ & 0.05$^{***}$ & 0.02$^{}$ & -0.03$^{***}$ & 0.02$^{}$ \\
 subregion\_N Africa & -0.12$^{***}$ & -0.10$^{***}$ & -0.03$^{***}$ & 0.01$^{*}$ & -0.06$^{***}$ \\
 subregion\_N America & -0.05$^{***}$ & -0.09$^{***}$ & -0.03$^{*}$ & 0.01$^{}$ & -0.04$^{***}$ \\
 subregion\_N Europe & 0.01$^{}$ & 0.03$^{***}$ & 0.01$^{}$ & -0.02$^{***}$ & 0.01$^{}$ \\
 subregion\_Oceania & -0.05$^{***}$ & -0.01$^{}$ & -0.05$^{***}$ & -0.03$^{***}$ & -0.04$^{***}$ \\
 subregion\_S Asia & -0.13$^{***}$ & -0.16$^{***}$ & -0.07$^{***}$ & 0.01$^{}$ & -0.09$^{***}$ \\
 subregion\_S Europe & -0.05$^{***}$ & -0.04$^{***}$ & -0.03$^{**}$ & -0.03$^{***}$ & -0.04$^{***}$ \\
 subregion\_SE Asia & -0.19$^{***}$ & -0.15$^{***}$ & -0.08$^{***}$ & -0.00$^{}$ & -0.11$^{***}$ \\
 subregion\_Sub-Saharan Africa & -0.03$^{}$ & -0.05$^{**}$ & 0.03$^{}$ & 0.03$^{***}$ & -0.00$^{}$ \\
 subregion\_W Asia & -0.12$^{***}$ & -0.12$^{***}$ & -0.06$^{***}$ & -0.03$^{***}$ & -0.08$^{***}$ \\
 subregion\_W Europe & -0.08$^{***}$ & -0.05$^{***}$ & -0.02$^{**}$ & -0.02$^{***}$ & -0.04$^{***}$ \\
\hline \\[-1.8ex]
 Observations & 296120 & 296122 & 296122 & 296122 & 296122 \\
 $R^2$ & 0.11 & 0.07 & 0.16 & 0.10 & 0.12 \\
 Adjusted $R^2$ & 0.11 & 0.07 & 0.16 & 0.10 & 0.12 \\
 Residual Std. Error & 0.60 & 0.77 & 0.64 & 0.36 & 0.46 \\
 F Statistic & 399.06$^{***}$ & 241.83$^{***}$ & 628.70$^{***}$ & 354.59$^{***}$ & 452.18$^{***}$ \\
\hline
\hline \\[-1.8ex]
\textit{Note:} & \multicolumn{5}{r}{$^{*}$p$<$0.05; $^{**}$p$<$0.01; $^{***}$p$<$0.001} \\
\end{tabular}
}
\caption{Coefficients for all regressions in \sref{sec:exp3}.
Stars represent p-values adjusted for multiple comparisons using a Bonferroni correction (* p<0.05, ** p<0.01, *** p<0.001).}
\label{tab:regression-coefficients-all}
\end{table*}

\begin{table*}[!htbp] \centering
\resizebox{\textwidth}{!}{
\begin{tabular}{lSSSSS}
\\[-1.8ex]\hline
\hline \\[-1.8ex]
\\[-1.8ex] & \multicolumn{1}{c}{Lexical} & \multicolumn{1}{c}{Semantic} & \multicolumn{1}{c}{Style} & \multicolumn{1}{c}{Syntactic} & \multicolumn{1}{c}{Overall}  \\
\hline \\[-1.8ex]
 topic\_General-short questions, story & -0.03$^{**}$ & 0.01$^{}$ & -0.02$^{}$ & -0.10$^{***}$ & -0.04$^{***}$ \\
 topic\_General-short requests & -0.49$^{***}$ & -0.53$^{***}$ & -0.24$^{***}$ & -0.09$^{**}$ & -0.34$^{***}$ \\
 topic\_Information-business & -0.03$^{}$ & -0.05$^{***}$ & -0.13$^{***}$ & -0.04$^{***}$ & -0.06$^{***}$ \\
 topic\_Information-chemistry & 0.02$^{}$ & -0.05$^{*}$ & -0.28$^{***}$ & -0.04$^{***}$ & -0.09$^{***}$ \\
 topic\_Information-history & 0.07$^{}$ & -0.03$^{}$ & 0.10$^{**}$ & 0.01$^{}$ & 0.04$^{}$ \\
 topic\_Information-language, programming & -0.15$^{***}$ & -0.21$^{***}$ & -0.11$^{***}$ & -0.17$^{***}$ & -0.16$^{***}$ \\
 topic\_Information-math, statistics & -0.05$^{***}$ & -0.10$^{***}$ & -0.39$^{***}$ & -0.11$^{***}$ & -0.16$^{***}$ \\
 topic\_Information-philosophy, physics & -0.14$^{***}$ & -0.13$^{***}$ & -0.22$^{***}$ & -0.14$^{***}$ & -0.16$^{***}$ \\
 topic\_Information-seo & -0.38$^{***}$ & -0.33$^{***}$ & -0.34$^{***}$ & -0.12$^{***}$ & -0.29$^{***}$ \\
 topic\_Jailbreak-crewbattles & -0.23$^{***}$ & -0.02$^{}$ & -0.16$^{***}$ & -0.09$^{***}$ & -0.13$^{***}$ \\
 topic\_Jailbreak-lucys, dan & -0.31$^{***}$ & -0.52$^{***}$ & -0.54$^{***}$ & -0.20$^{***}$ & -0.39$^{***}$ \\
 topic\_Jailbreak-math, code & -0.06$^{}$ & 0.01$^{}$ & -0.21$^{***}$ & -0.01$^{}$ & -0.07$^{**}$ \\
 topic\_Jailbreak-narotica & -0.16$^{***}$ & -0.16$^{***}$ & -0.10$^{***}$ & -0.16$^{***}$ & -0.14$^{***}$ \\
 topic\_Jailbreak-nsfwgpt & -0.13$^{***}$ & -0.11$^{***}$ & -0.19$^{***}$ & -0.16$^{***}$ & -0.15$^{***}$ \\
 topic\_Jailbreak-system & -0.16$^{***}$ & -0.11$^{***}$ & -0.26$^{***}$ & -0.12$^{***}$ & -0.16$^{***}$ \\
 topic\_Multilingual-japanese, chinese & -0.27$^{***}$ & -0.25$^{***}$ & -0.22$^{***}$ & 0.02$^{}$ & -0.18$^{***}$ \\
 topic\_Multilingual-russian, chinese & -0.33$^{***}$ & -0.11$^{}$ & -0.33$^{***}$ & -0.10$^{***}$ & -0.22$^{***}$ \\
 topic\_Programming-agent setup1 & -0.30$^{***}$ & -0.56$^{***}$ & -0.30$^{***}$ & -0.13$^{***}$ & -0.32$^{***}$ \\
 topic\_Programming-agent setup2 & -0.22$^{***}$ & -0.16$^{***}$ & -0.16$^{***}$ & -0.06$^{***}$ & -0.15$^{***}$ \\
 topic\_Programming-audio, math & -0.09$^{***}$ & -0.15$^{***}$ & 0.05$^{}$ & 0.05$^{*}$ & -0.04$^{}$ \\
 topic\_Programming-data science & -0.26$^{***}$ & -0.28$^{***}$ & -0.55$^{***}$ & -0.19$^{***}$ & -0.32$^{***}$ \\
 topic\_Programming-front end & -0.28$^{***}$ & -0.33$^{***}$ & -0.50$^{***}$ & -0.25$^{***}$ & -0.34$^{***}$ \\
 topic\_Programming-java & -0.34$^{***}$ & -0.45$^{***}$ & -0.61$^{***}$ & -0.16$^{***}$ & -0.39$^{***}$ \\
 topic\_Programming-java, app & -0.25$^{***}$ & -0.31$^{***}$ & -0.41$^{***}$ & -0.12$^{***}$ & -0.27$^{***}$ \\
 topic\_Programming-python, data science & -0.20$^{***}$ & -0.24$^{***}$ & -0.45$^{***}$ & -0.19$^{***}$ & -0.27$^{***}$ \\
 topic\_Roleplay setup-sexual & -0.11$^{***}$ & -0.08$^{***}$ & -0.08$^{***}$ & -0.27$^{***}$ & -0.13$^{***}$ \\
 topic\_Roleplay setup-teen drama & 0.03$^{}$ & -0.02$^{}$ & 0.10$^{***}$ & -0.08$^{***}$ & 0.01$^{}$ \\
 topic\_Roleplay setup-transmorph, sexual & -0.06$^{***}$ & -0.05$^{**}$ & 0.13$^{***}$ & -0.06$^{***}$ & -0.01$^{}$ \\
 topic\_Story, Programming-sci-fi, svg image & -0.04$^{}$ & -0.06$^{}$ & -0.14$^{***}$ & -0.14$^{***}$ & -0.09$^{***}$ \\
 topic\_Story-alex-zane, anime & 0.04$^{}$ & 0.14$^{***}$ & 0.22$^{***}$ & -0.18$^{***}$ & 0.05$^{***}$ \\
 topic\_Story-alternative history & 0.05$^{***}$ & 0.06$^{***}$ & -0.00$^{}$ & -0.04$^{***}$ & 0.02$^{}$ \\
 topic\_Story-animal, monster & 0.10$^{**}$ & 0.22$^{***}$ & 0.13$^{***}$ & 0.03$^{}$ & 0.12$^{***}$ \\
 topic\_Story-bathroom, sexual & -0.00$^{}$ & 0.14$^{*}$ & 0.00$^{}$ & 0.09$^{***}$ & 0.06$^{}$ \\
 topic\_Story-boyband & -0.24$^{***}$ & -0.14$^{***}$ & -0.17$^{***}$ & -0.07$^{***}$ & -0.15$^{***}$ \\
 topic\_Story-comedy1 & -0.06$^{}$ & -0.11$^{**}$ & -0.02$^{}$ & -0.20$^{***}$ & -0.10$^{***}$ \\
 topic\_Story-comedy2 & -0.07$^{***}$ & -0.01$^{}$ & 0.05$^{**}$ & -0.10$^{***}$ & -0.03$^{*}$ \\
 topic\_Story-fan fiction & 0.17$^{***}$ & 0.21$^{***}$ & 0.12$^{***}$ & -0.02$^{}$ & 0.12$^{***}$ \\
 topic\_Story-japanese musician & 0.03$^{}$ & -0.10$^{}$ & 0.02$^{}$ & -0.04$^{}$ & -0.02$^{}$ \\
 topic\_Story-kid, girl & -0.00$^{}$ & -0.10$^{***}$ & -0.07$^{***}$ & -0.09$^{***}$ & -0.07$^{***}$ \\
 topic\_Story-kids' show & -0.08$^{**}$ & -0.08$^{*}$ & -0.11$^{***}$ & -0.11$^{***}$ & -0.09$^{***}$ \\
 topic\_Story-literature club & 0.03$^{}$ & 0.06$^{}$ & -0.05$^{}$ & -0.12$^{***}$ & -0.02$^{}$ \\
 topic\_Story-movies & -0.05$^{**}$ & -0.03$^{}$ & 0.03$^{}$ & -0.08$^{***}$ & -0.03$^{*}$ \\
 topic\_Story-pokemon, casual & -0.01$^{}$ & 0.05$^{}$ & -0.09$^{}$ & 0.01$^{}$ & -0.01$^{}$ \\
 topic\_Story-robot & -0.18$^{}$ & -0.06$^{}$ & -0.29$^{***}$ & -0.03$^{}$ & -0.14$^{}$ \\
 topic\_Story-sci-fi, magic & 0.01$^{}$ & -0.04$^{}$ & 0.05$^{***}$ & -0.14$^{***}$ & -0.03$^{*}$ \\
 topic\_Story-superhero & 0.03$^{}$ & -0.40$^{***}$ & 0.29$^{***}$ & -0.23$^{***}$ & -0.08$^{***}$ \\
 topic\_Text-to-Image prompt-human & -0.19$^{***}$ & -0.26$^{***}$ & -0.16$^{***}$ & -0.09$^{***}$ & -0.17$^{***}$ \\
 topic\_Text-to-Image prompt-scene1 & -0.33$^{***}$ & -0.34$^{***}$ & -0.25$^{***}$ & -0.16$^{***}$ & -0.27$^{***}$ \\
 topic\_Text-to-Image prompt-scene2 & -0.24$^{***}$ & -0.25$^{***}$ & -0.20$^{***}$ & -0.01$^{}$ & -0.18$^{***}$ \\
\hline
\hline \\[-1.8ex]
\textit{Note:} & \multicolumn{5}{r}{$^{*}$p$<$0.05; $^{**}$p$<$0.01; $^{***}$p$<$0.001} \\
\end{tabular}
}
\caption{Continuation of Table \ref{tab:regression-coefficients-all}}
\label{tab:regression-coefficients-all-continued}
\end{table*}

\end{document}